%% file: main.tex
\documentclass{article} 

\PassOptionsToPackage{numbers}{natbib}

\usepackage[final]{neurips_2024}

\usepackage[table,xcdraw]{xcolor}
\usepackage{hyperref}
\usepackage{url}
\usepackage{enumitem}
\usepackage{graphicx}
\usepackage{xcolor}
\usepackage{booktabs}
\usepackage{multirow}
\usepackage{amsmath}
\usepackage{bbm}
\usepackage{amsfonts}
\usepackage{amssymb}
\usepackage{tcolorbox}
\usepackage{multirow}
\usepackage{subcaption}
\usepackage{listings}
\usepackage{inconsolata}
\usepackage{dsfont}
\usepackage{bbold}
\usepackage{mathbbol}

\definecolor{mygreen}{HTML}{008000}
\definecolor{olive}{HTML}{baba65}
\definecolor{beige}{HTML}{d4b996}

\definecolor{syntaxerror}{rgb}{0.890, 0.102, 0.110}
\definecolor{wrongentrelpropertytype}{rgb}{0.651, 0.808, 0.890}
\definecolor{reverseddirection}{rgb}{0.122, 0.471, 0.706}
\definecolor{entitylinking}{rgb}{0.698, 0.875, 0.541}
\definecolor{schemaviolation}{rgb}{0.984, 0.604, 0.600}
\definecolor{patternnotalignedwithquestion}{rgb}{0.992, 0.749, 0.435}
\definecolor{incorrectgrouping}{rgb}{0.792, 0.698, 0.839}
\definecolor{incorrectfiltering}{rgb}{0.416, 0.239, 0.604}
\definecolor{resultsnotalignedwithquestion}{rgb}{0.9, 0.9, 0.500}
\definecolor{incorrectdeduplication}{rgb}{0.694, 0.349, 0.157}

\lstdefinestyle{TinySQL}{
    language=SQL, 
    morekeywords={
        WITH
    }, 
    basicstyle=\ttfamily\tiny, 
    keywordstyle=\color{mygreen}\bfseries, 
    stringstyle=\color{purple}, 
    commentstyle=\color{gray}\itshape, 
    numbers=none, 
    numberstyle=\tiny\color{gray}, 
    stepnumber=1, 
    numbersep=10pt, 
    tabsize=2, 
    captionpos=b, 
    breaklines=true, 
    breakatwhitespace=true, 
    showspaces=false, 
    showstringspaces=false, 
    escapeinside={(*@}{@*)}, 
    frame=none, 
    aboveskip=0pt, 
    belowskip=0pt, 
    extendedchars=true,
    literate={á}{{\'a}}1 {ã}{{\~a}}1 {é}{{\'e}}1 {í}{{\'i}}1 {ó}{{\'o}}1 {ú}{{\'u}}1 {ç}{{\c{c}}}1,
}

\lstdefinestyle{TinyJSON}{
    language=,
    basicstyle=\ttfamily\tiny, 
    keywordstyle=\color{mygreen}\bfseries, 
    stringstyle=\color{purple}, 
    commentstyle=\color{gray}\itshape, 
    numbers=none, 
    numberstyle=\tiny\color{gray}, 
    stepnumber=1, 
    numbersep=10pt, 
    tabsize=2, 
    captionpos=b, 
    breaklines=true, 
    breakatwhitespace=true, 
    showspaces=false, 
    showstringspaces=false, 
    escapeinside={(*@}{@*)}, 
    frame=none, 
    aboveskip=0pt, 
    belowskip=0pt, 
    extendedchars=true,
    literate={á}{{\'a}}1 {ã}{{\~a}}1 {é}{{\'e}}1 {í}{{\'i}}1 {ó}{{\'o}}1 {ú}{{\'u}}1 {ç}{{\c{c}}}1,
}

\input{macros.tex}

\definecolor{brickred}{RGB}{110, 30, 28}
\definecolor{brickred}{RGB}{233, 158, 122}
\definecolor{brickred}{RGB}{217, 119, 87}
\definecolor{mypurple}{RGB}{145, 112, 224}

\title{%
  \textit{ARCS}: Towards Precise Text-to-SQL via\\
  Structured Disambiguation
}

\author{%
 Yihao Hu\thanks{The authors contributed equally to this work. This work was done during Yihao's internship at Megagon Labs. Correspondence to \texttt{yanlin@megagon.ai}.}~~\textcolor{brickred}{$^\alpha$} \quad Yanlin Feng$^{*}$\textcolor{mypurple}{$^{\beta}$} \quad Naoki Otani\textcolor{mypurple}{$^\beta$} \quad Nikita Bhutani\textcolor{mypurple}{$^\beta$} \vspace{0.3em} \\ 
  \textcolor{brickred}{$^\alpha$}Databricks \ \ \textcolor{mypurple}{$^\beta$}Megagon Labs\\
  \texttt{yihao.hu@databricks.com}  \ \ \texttt{\{yanlin, naoki, nikita\}@megagon.ai} \vspace{1.2em}\\
\raisebox{-1.5pt}{\includegraphics[height=1.05em]{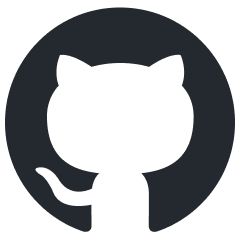}} ~\raisebox{-1.6pt}{\includegraphics[height=1.2em]{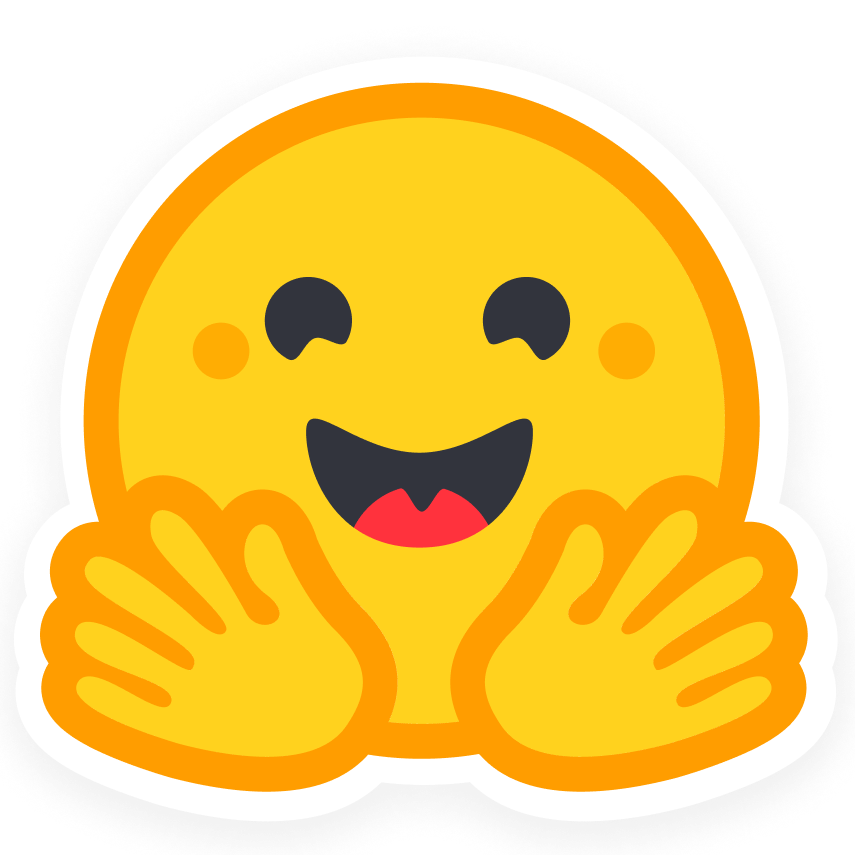}} ~\href{https://megagonlabs.github.io/tabulaflow/research/arcs/}
     {\textcolor{cyan}{\texttt{megagonlabs.github.io/tabulaflow/research/arcs}}}
}
\begin{document}

\maketitle


\begin{abstract}

As text-to-SQL systems move beyond demonstrations toward real-world deployment, ambiguity in user questions becomes a primary source of errors. Such ambiguities are often subtle, domain- or data-specific, and can silently cause system outputs to deviate from the user's true intent. Ambiguity is traditionally addressed through conversational clarification, which is often inefficient, cognitively demanding, and poorly aligned with real-world user workflows. We propose \textbf{structured disambiguation}, a new paradigm in which ambiguity is resolved through explicit, constrained interactions rather than free-form dialogue.  We construct \textbf{ARCS} (\textbf{A}mbiguity \textbf{R}esolution \textbf{C}orpus for \textbf{S}QL), the first text-to-SQL benchmark featuring naturally occurring, unconstrained ambiguities over real-world databases, with complete annotations of all valid ambiguity points, interpretations, and SQL queries. Experimental results show that text-to-SQL remains challenging in the presence of ambiguity: \texttt{gpt-6-sol} achieves only 51\% end-to-end execution accuracy, and no open-source model exceeds 27\%.

\end{abstract}

\input{sections/1-intro}

\input{sections/2-disambiguation}

\input{sections/3-taxonomy}

\input{sections/4-benchmark-construction}

\input{sections/5-experiments}

\input{sections/7-conclusion}

\bibliography{iclr2026_conference}
\bibliographystyle{iclr2026_conference}

\clearpage

\appendix

\input{sections/appendix}

\end{document}

%% file: macros.tex
\usepackage{xspace}
\usepackage{xcolor}

\definecolor{naokigreen}{RGB}{0,170,0} 

\newcommand{\eg}{e.g.\@}

%% file: sections/1-intro.tex
\section{Introduction}\label{sec:intro}


\input{figures/structured_disamb}

As state-of-the-art performance on Spider 2.0~\citep{lei2024spider2} surpasses 80\%, this milestone marks a notable advancement in recent text-to-SQL systems~\citep{pourreza_reasoning-sql_2025,pourreza_chase-sql_2024,ali_state---art_2025,shkapenyuk_automatic_2025,wang_agentar-scale-sql_2025,deng_reforce_2025,hao_text--sql_2025,li2025deepeye,zhang2025reward}, which serve as a bridge between enterprise databases and generative AI. However, most widely used benchmarks~\cite{zhongSeq2SQL2017,Yu&al.18c,li2024can,lei2024spider2,chen2024beaver,lee-etal-2021-kaggledbqa} consist of unambiguous, lengthy, often over-specified questions, which diverge from how user ask questions in practice. This reveals a less discussed issue behind strong benchmark performance: the ambiguity inherent in natural language that can cause system outputs to deviate from the true user intent~\citep{wang-etal-2023-know,bhaskar-etal-2023-benchmarking,DBLP:conf/nips/SaparinaL24}.

For example, consider the question ``How many GitHub issues are opened in 2025?'': should this count only newly created issues, or should it also include issues reopened during 2025? Similarly, the question ``What was the total sales revenue?'' is ambiguous as ``sales revenue'' can be calculated differently based on refunded, canceled, or partially returned orders. Such ambiguities are often subtle and highly specific to domains and data. When data produced by text-to-SQL systems is used to support critical business decisions (\eg, setting annual budgets based on retrieved revenue data), these ambiguities can have a substantial business impact.


Traditional conversational disambiguation~\citep{dong-etal-2025-practiq,birdinteract2025} relies on the user to read, recall, and articulate specific constraints in natural language. In contrast, we propose structured disambiguation, where users simply recognize the correct intent through multiple-choice selections (when interpretations are finite) and value adjustment (for numeric thresholds). This approach leverages the HCI principle of ``recognition rather than recall''~\cite{nielsen-1994-enhancing}, thereby achieving more efficient disambiguation with reduced cognitive load. Furthermore, structured disambiguation improves the workflow in several ways: (1) the SQL query and its resulting table are generated immediately upon selection, and (2) users can easily compare the results of different resolutions and re-select among them. An illustrative comparison of conversational disambiguation and structured disambiguation is shown in \autoref{fig:structured_disamb}. Structured disambiguation substantially reduces user effort and enables more interactive disambiguation.


To enable rigorous evaluation of text-to-SQL systems under realistic ambiguous user questions, we introduce ARCS: \textbf{A}mbiguity \textbf{R}esolution \textbf{C}orpus for \textbf{S}QL. When constructing ARCS, we adopt a fundamentally different methodology from prior work~\citep{wang-etal-2023-know,bhaskar-etal-2023-benchmarking,DBLP:conf/nips/SaparinaL24}. Rather than predefining a small set of ambiguity types and artificially injecting them into questions and databases, we identify intrinsic ambiguities within realistic questions based on unmodified, real-world databases. This process yields questions naturally containing multiple, interacting ambiguity phrases. In addition, ARCS supports end-to-end evaluation through a user simulator, without imposing constraints on the system architecture or the disambiguation strategy. The benchmark offers complete annotations of all valid ambiguity points, interpretations, and SQL queries, enabling disentangling disambiguation performance from SQL generation performance.

Our experimental results on ARCS reveal several interesting findings. When questions are ambiguous, Text-to-SQL remains a significant challenge for modern large language models (LLMs): \texttt{gpt-6-sol} achieves only 51\% end-to-end execution accuracy, and no open-source LLM exceeds 27\%. Furthermore, frontier LLMs from different providers, while having similar text-to-SQL capabilities, exhibit very different disambiguation performance, suggesting that disambiguation may be an underrepresented objective during LLM training.

In summary, our work makes the following main contributions:

\begin{itemize}[left=10pt]
\item Structured disambiguation, a new disambiguation interface for text-to-SQL.
\item ARCS, a reliable and versatile benchmark with 311 realistic disambiguation tasks.
\item A taxonomy for systematically categorizing ambiguities in text-to-SQL.
\item Extensive experiments and analysis of current state-of-the-art LLMs.
\end{itemize}

%% file: figures/structured_disamb.tex
\begin{figure*}
    \centering
    \includegraphics[width=\linewidth]{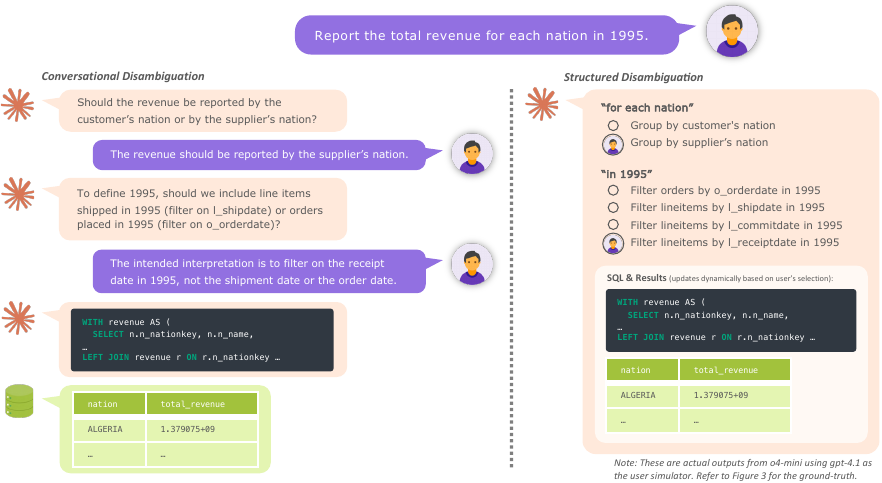}
    \caption{Comparison of conversational disambiguation and structured disambiguation. Structured disambiguation allows users to compare and select interpretations for multiple ambiguity points based on the SQL queries and execution results.}
    \label{fig:structured_disamb}
\end{figure*}

%% file: sections/2-disambiguation.tex
\section{Structured Disambiguation for Text-to-SQL}\label{sec:disambiguation}


Ambiguity in natural language is often open-ended and admits a large number of possible interpretations. Consequently, the default approach to disambiguation across many NLP tasks is to ask clarification questions and rely on users to provide additional information in natural language~\citep{dong-etal-2025-practiq,niwa2024ambignlg,Zhang2024ModelingFC,lee-etal-2023-asking}. However, when a question is contextualized within a database, the space of valid interpretations is often substantially reduced because a database contains only a finite set of schema elements (\eg, tables and columns). This makes it possible to enumerate all possible interpretations and transforms the user's task from an open-ended recall problem (answering ``What do you mean?'') to a constrained recognition problem (selecting ``A or B'')~\citep{nielsen-1994-enhancing} for more efficient disambiguation.


Recent work has begun to explore this direction for Text-to-SQL. For example, \cite{saparina-lapata-2025-disambiguate} propose generating all possible interpretations for a given question. However, their approach remains incomplete in two key aspects. First, the enumeration is unstructured; when multiple ambiguity points are present, listing every full-sentence interpretations causes the number of candidates to grow exponentially. Second, it cannot handle vagueness, where the space of interpretations is unbounded and cannot be enumerated as a fixed list (\eg ``high transaction amount'').\footnote{Although vagueness is often distinguished from ambiguity~\cite{zwicky-and-sadock-1975-ambiguity}, we follow \cite{saparina-lapata-2025-disambiguate} and use \textit{ambiguity} to include vagueness in our work for brevity.}

Structured disambiguation addresses these limitations by decomposing ambiguity and resolving each ambiguity point separately. Our framework consists of three phases:

\paragraph{Disambiguation}
Given a question, we first identify a set of ambiguity points or phrases, denoted by 
\[
\mathcal{A} = \mathcal{A}_{\text{finite}} \cup \mathcal{A}_{\text{infinite}} .
\]
Here, \(\mathcal{A}_{\text{finite}}\) denotes the set of ambiguity points with a finite number of possible interpretations. For each ambiguity point \(a \in \mathcal{A}_{\text{finite}}\), we identify all valid interpretations \(I_a\). \(\mathcal{A}_{\text{infinite}}\) denotes ambiguity points with an unbounded interpretation space (\eg, numeric thresholds). For each \(a \in \mathcal{A}_{\text{infinite}}\), we introduce a parameter variable \(v_a\).

\paragraph{User Interaction}
After disambiguation, we present the user with an interaction panel consisting of \(|\mathcal{A}_{\text{finite}}|\) multiple-choice questions and \(|\mathcal{A}_{\text{infinite}}|\) value sliders. For each \(a \in \mathcal{A}_{\text{finite}}\), the user resolves the ambiguity by selecting the intended interpretation. For each \(a \in \mathcal{A}_{\text{infinite}}\), the user resolves the ambiguity by selecting a value via the slider and choosing the associated operator (\eg, \(>\) vs.\ \(\geq\)). For both types of ambiguity points, the user may also choose to reject the ambiguity (\eg, when it is incorrect or can be inferred from the database context).

\paragraph{SQL Generation}
Let \(\mathcal{A}_{\text{finite}}^{+} \subseteq \mathcal{A}_{\text{finite}}\) denote the set of finite ambiguity points not rejected by the user, and let \(\mathcal{I}^{*}\) denote the interpretation combination selected by the user, where
\[
\mathcal{I}^{*} \in \prod_{a \in \mathcal{A}_{\text{finite}}^{+}} I_a .
\]
We generate a parameterized SQL query corresponding to \(\mathcal{I}^{*}\), with variables \(\{v_a \mid a \in \mathcal{A}_{\text{infinite}}^{+}\}\). When executing the query to retrieve the final result table, these variables are instantiated using the values and operators specified by the user during interaction. The SQL generation step can be repeated on demand when the user selects a different interpretation combination. To further reduce latency, a caching mechanism can be employed by pre-generating SQL queries for all interpretation combinations \(\mathcal{I}\) in parallel prior to user interaction.

Importantly, the disambiguation and SQL generation components are implementation-agnostic. In particular, any text-to-SQL system can be substituted as the underlying SQL generation module. Details of our concrete implementation are provided in \autoref{appsec:technical}.

%% file: sections/3-taxonomy.tex
\input{figures/taxonomy}


\section{Categorizing Ambiguity in Text-to-SQL}\label{sec:taxonomy}


The first step toward building a benchmark with broad coverage of diverse ambiguity types is to develop a taxonomy that systematically categorizes the ambiguity space. A key challenge in characterizing ambiguity in text-to-SQL is that it is inherently \textit{intersectional:} it arises from the friction between natural language and a specific database schema. For instance, a linguistically ambiguous phrase (\eg, ``New York'' referring to either New York City or New York State) may be unambiguous given a specific schema (\eg, the database only has a city column), while a linguistically precise phrase may become ambiguous due to schema redundancy (\eg, multiple address-related columns encode city information). We address this intersectional nature by proposing a taxonomy with two orthogonal dimensions: a \emph{linguistic dimension}, which captures the linguistic source of the ambiguity, and a \emph{database dimension}, which captures how the ambiguity maps to database elements. Both dimensions are exhaustive by design. \autoref{fig:taxonomy} provides an illustration of our taxonomy.

\paragraph{Linguistic dimension}
Along the linguistic dimension, we categorize ambiguity into the following two types:

\emph{Semantic ambiguity.}
Ambiguity arising from multiple possible meanings of words or phrases (\eg, polysemy and vague predicates).

\emph{Syntactic ambiguity.}
Ambiguity arising from multiple possible syntactic structures in a sentence (\eg, modifier attachment).

\paragraph{Database dimension}
Along the database dimension, we categorize ambiguity into four types\footnote{Unlike the linguistic dimension, distinctions between categories in the database dimension are sometimes subtle. In such cases, we determine the ambiguity type based on the resulting SQL queries. For example, we classify a phrase as exhibiting column ambiguity if the only differences among its valid SQL queries lie in the selected column and column-dependent logic.}:

\emph{Column ambiguity.}
An ambiguous phrase can map to multiple columns in the database schema, such as attributes with similar semantics across columns.

\emph{Table ambiguity.}
An ambiguous phrase can map to multiple tables in the database schema, for example when the same attribute appears in different tables, leading to alternative table selections or join paths.

\emph{Value ambiguity.}
An ambiguous phrase can map to multiple possible values within a column, including categorical values, numeric thresholds, or entities that satisfy the same description.

\emph{Computation ambiguity.}
An ambiguous phrase can map to multiple computation strategies, such as different aggregation functions, logical operators, or arithmetic expressions.


Unlike our approach, prior work focused on special cases along either the linguistic dimension~\citep{DBLP:conf/nips/SaparinaL24} or the database dimension~\citep{wang-etal-2023-know,bhaskar-etal-2023-benchmarking} in isolation, or conflate the two~\citep{birdinteract2025}. By treating these dimensions as orthogonal, our taxonomy provides an opportunity for diagnosing whether a system fails at understanding language nuances or at grounding them to database elements. This separation is practically useful: it enables developers to distinguish between ``parsing errors'' (linguistic) and ``schema alignment errors'' (database) for targeted improvements in system reliability.

%% file: figures/taxonomy.tex
\begin{figure*}[t]
    \centering
    \includegraphics[width=\linewidth]{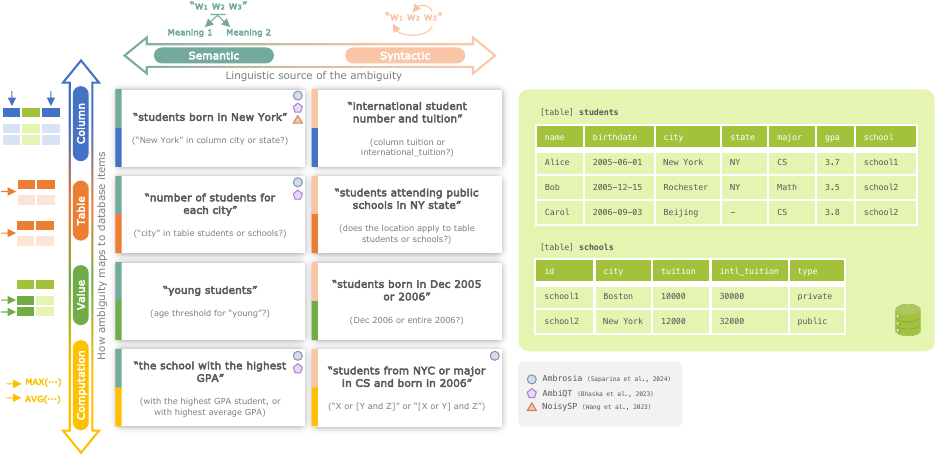}
    \caption{Illustration of the ARCS taxonomy on a curated database.}
    \label{fig:taxonomy}
\end{figure*}

%% file: sections/4-benchmark-construction.tex
\input{figures/examples} 

\section{Benchmark Construction} \label{sec:benchmark}

To evaluate structured disambiguation in text-to-SQL, we construct ARCS, a benchmark dataset that covers the full spectrum of the ambiguity taxonomy introduced in \S\ref{sec:taxonomy} and provides complete annotations of all ambiguity points, all valid interpretations, and their corresponding SQL queries (see \autoref{fig:examples}). As illustrated in \autoref{fig:evaluation}, ARCS supports two complementary evaluation settings:
\paragraph{End-to-End Setting} The system receives a natural language question and may freely interact with the database and user simulator in any order before generating the final SQL query. This setting imposes no constraints on agent architecture.
\paragraph{Fine-Grained Setting} This setting disentangles the pipeline into two phases. In the disambiguation phase, the system must output the complete set of ambiguity points and valid interpretations given the question and database. In the SQL generation phase, the system generates the final SQL query given the ground-truth disambiguation.

\subsection{Database Construction}\label{subsec:db-construction}

We prioritize two considerations when constructing database instances. First, the databases should be realistic, of significant real-world interest and, when available, based on real data with sufficient coverage and scale. Second, the databases should be well suited for creating ambiguous questions (\eg, through semantically overlapping columns), and collectively they should support all eight ambiguity types. Given these considerations, we select existing databases and use them as-is, rather than synthesizing small-scale databases or modifying existing ones (\eg, by adding columns with synthetic data), as done in prior work~\citep{DBLP:conf/nips/SaparinaL24,wang-etal-2023-know,bhaskar-etal-2023-benchmarking}.

In total, we select five databases from BIRD~\citep{li2024can} and one database from Spider 2.0~\citep{lei2024spider2}. All six databases are converted to SQLite to enable convenient local access for researchers, with a total size of 8.3 GB. The selected databases cover a diverse range of domains, including banking, retails, student activities, NBA, Stack Exchange forum, and GitHub repositories.

\subsection{Ambiguous Question Generation}

To generate a large pool of candidate questions at scale, we use \texttt{gpt-4.1}, conditioned on the target database schema, our taxonomy, and a small set of human-curated exemplar questions. We provide the taxonomy both as illustrative examples and as labeling guidelines for ambiguity points; however, the model is not constrained to replicate the ambiguity shown in the taxonomy and is free to generate novel ambiguities. The model is prompted to produce a set of ambiguous questions, and this process is run repeatedly with the temperature set to 0.7 until the desired number of questions is reached. We generate 180 questions for each database.

\subsection{Ambiguity and SQL Annotation} \label{subsec:annotation}

After the initial generation, human annotators review all candidate questions and select those that are both realistic and genuinely ambiguous. For each selected question, annotators then identify the ambiguity points and enumerate all valid interpretations associated with each finite ambiguity point\footnote{During question generation, the model is additionally prompted to produce an initial annotation of ambiguity points, interpretations and SQL queries, which serves as a starting point for subsequent human annotation.}. Throughout this process, annotators may revise the question text and add or remove ambiguity points as needed.

One issue overlooked in prior work~\citep{birdinteract2025} is the distinction between true ambiguity and global annotation conventions. For example, choices such as whether to return ties when computing an argmax, or whether to round numeric values reflect dataset-level conventions rather than inherent per-instance ambiguity. Treating these as ambiguities would be inappropriate, as they would trivially affect nearly all questions. In ARCS, we provide an explicit specification of all annotation conventions that can be embedded in the LLM prompt, so that the model adheres to these conventions instead of interpreting them as ambiguities. This setup is analogous to global user preference settings that can be configured in real-world applications.

Following the notation in \S\ref{sec:disambiguation}, let \( \mathcal{A}_{\text{finite}}^{*} \) denote the set of annotated finite ambiguity points. For each interpretation combination
\( \mathcal{I} \in \prod_{a \in \mathcal{A}_{\text{finite}}^{*}} I_a \),
human annotators produce a parameterized SQL query with variables
\( \{ v_a \mid a \in \mathcal{A}_{\text{infinite}}^{*} \} \). 

Due to the need for high expertise in both linguistics and databases, the authors carried out all annotations to ensure the highest level of annotation quality. After an initial round of annotation, the authors conducted a second round using the finalized annotation guidelines, as well as a final verification pass. Additional details are provided in \autoref{appsec:benchmark_details}. On average, annotating a single question requires 2.5 hours of effort by one annotator.



\subsection{User Simulator} \label{subsec:user_simulator}

To support the evaluation of arbitrary disambiguation and user interaction methods, we design an LLM-based user simulator that exposes an \texttt{ask\_user} interface. The \texttt{ask\_user} interface supports free-text questions, multiple-choice questions, and value-selection questions. When replaced with a real user in a frontend, these interaction types can be realized as text input fields, multiple-choice selectors, and value sliders, respectively. During inference, the system invokes a user simulator instance as needed, providing full flexibility in deciding when and how to interact with the user.

The user simulator has access the ground-truth resolution for each ambiguity point that leads to one correct SQL query per question. For finite ambiguity points, we randomly sample one interpretation from \(I_a\). For infinite ambiguity points, a human annotator specifies a concrete value and an operator such that the resulting SQL query yields a non-empty result. We enforce that the final user-intended SQL query always produces a non-empty output. We perform multiple samplings per question, resulting in 331 task instances sampled from 101 unique questions, with 413 unique SQL queries in total, a scale comparable to recent text-to-SQL benchmarks~\citep{lei2024spider2, chen2024beaver}. Multiple sampling is necessary to reduce bias toward an easy or difficult interpretation in the end-to-end setting, and 331 instances provide a sweet spot that balances experimental cost and reliability.

In our initial exploration, we found that a naive implementation where all annotated ambiguity points and their intended interpretations are provided in the system prompt does not work as expected due to information leakage. Specifically, the LLM often reveals information related to other ambiguity points even when they are not explicitly queried, a phenomenon previously observed in prior work~\citep{birdinteract2025}. To mitigate this issue, we adopt a two-stage user simulator following this prior work: the LLM first identifies the ambiguity points relevant to the question and then uses only those identified ambiguity points to generate its response. The prompts for user simulator are provided in \autoref{appsec:technical}.

\input{figures/evaluation}

\subsection{Evaluation Metrics}
As described earlier, our benchmark supports two evaluation settings: \emph{end-to-end evaluation} and \emph{fine-grained evaluation}, as illustrated in \autoref{fig:evaluation}. All metrics are computed at the task instance level and aggregated using a simple average over the entire benchmark.

\subsubsection{End-to-End Evaluation}
The primary metrics for end-to-end evaluation are as follows:
\begin{itemize}[left=10pt]
    \item \textbf{EX}$_{\text{e2e}}$: end-to-end execution accuracy that measures whether the execution result of the final predicted SQL query matches that of the ground-truth user-intended SQL query.
    \item \textbf{User Effort}: an approximation of the interaction cost, computed as $0.1 \times$ (number of input tokens) plus the number of output tokens produced by the user simulator.
    \item \textbf{Cost}: the monetary cost per task instance, excluding the cost of the user simulator.
\end{itemize}

An important design choice in computing execution accuracy is that we allow the predicted SQL query to return additional columns beyond those in the ground truth. This is particularly important for ambiguous questions, where a highly intelligent LLM may choose not to explicitly disambiguate and instead directly compute a result table that covers all valid interpretations (e.g., returning both gross revenue and net revenue in a single table). In such cases, the prediction receives $\mathrm{EX}_{\text{e2e}} = 1$ and a user effort of 0.

\input{tables/main_result}

\subsubsection{Fine-Grained Evaluation}
Fine-grained evaluation measures performance of disambiguation and text-to-SQL generation in isolation. The primary metrics are:
\begin{itemize}[left=10pt]
    \item \textbf{Full Recall}: measures disambiguation performance by checking whether \emph{all} ground-truth ambiguity points and their corresponding interpretations are identified.
    \item \textbf{Perfect}: measures disambiguation performance by checking whether \emph{exactly} all and only the ground-truth ambiguity points and interpretations are identified.
    \item \textbf{EX}$_{\text{disambiguated}}$: text-to-SQL generation execution accuracy given ground-truth disambiguation.
\end{itemize}
{Full Recall} and {Perfect} are computed by aligning the predicted ambiguity points and their interpretations with the annotated ones. We have conducted experiments to ensure that the alignment is highly deterministic and consistent with human judgment. Additional details are provided in \autoref{appsec:technical}.

%% file: figures/examples.tex
\begin{figure*}
    \centering
    \includegraphics[width=\linewidth]{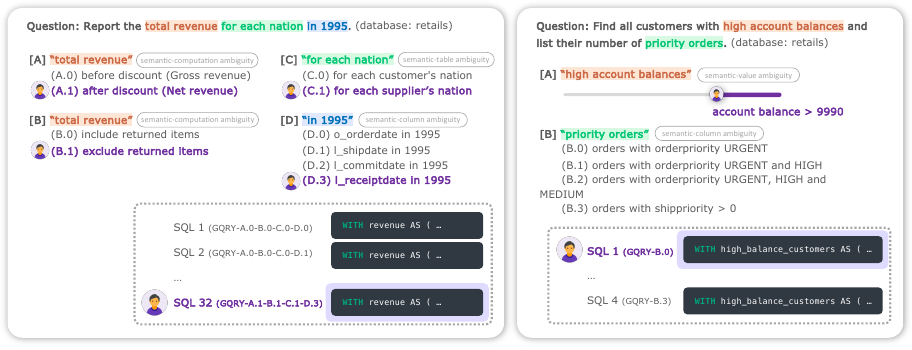}
    \caption{Two task instances from ARCS with annotations of all valid ambiguity points, interpretations, and SQL queries.}
    \label{fig:examples}
\end{figure*}

%% file: figures/evaluation.tex
\begin{figure*}
    \centering
    \includegraphics[width=\linewidth]{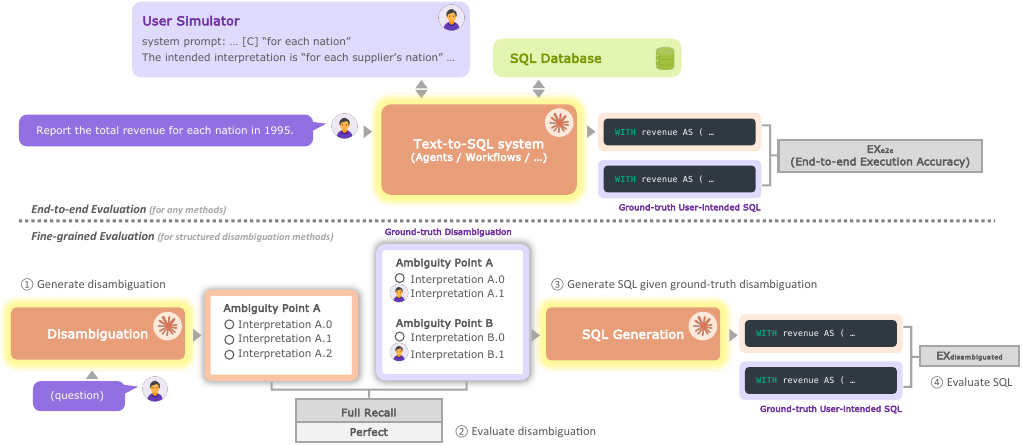}
    \caption{Evaluation settings and metrics for ARCS.}
    \label{fig:evaluation}
\end{figure*}


%% file: tables/main_result.tex
\begin{table*}[ht]
\scriptsize
\centering
\scalebox{1.0}{
\begin{tabular}{l r r r r r r}
\toprule
& \multicolumn{2}{c}{\textbf{Disambiguation}} &
\multicolumn{1}{c}{\textbf{SQL Generation}} &
\multicolumn{3}{c}{\textbf{End-to-end}}
 \\
\cmidrule(lr){2-3} \cmidrule(lr){4-4} \cmidrule(lr){5-7}
\textbf{Method} & \textbf{Full Recall} (\%) & \textbf{Perfect} (\%) & \textbf{EX}$_\text{disambiguated}$ (\%) & \textbf{EX}$_\text{e2e}$  (\%) & \textbf{$\Delta$EX} & \textbf{Cost} (\$) \\

\midrule
\rowcolor[gray]{0.9}
\multicolumn{7}{l}{\textit{Open-source LLMs}} \vspace{2pt} \\
\texttt{gptoss-20b}                    & 4.18 & 0.96 & 27.97 & 7.40 & -20.57 & 0.0003 \\
\texttt{gptoss-120b}                   & 16.72 & 1.61 & 52.09 & 26.05 & -26.04 & 0.003 \\
\texttt{qwen3-8b}                      & 6.11 & 0.00 & 11.25 & 5.79 & -5.46 & 0.009 \\
\texttt{qwen3-235b-a22b-instruct-2507} & 6.43 & 2.57 & 37.30 & 20.26 & -17.04 & 0.02 \\
\texttt{qwen3-coder-480b}              & 6.11 & 2.57 & 26.69 & 15.76 & -10.93 & 0.03 \\
\texttt{deepseek-v3.1}                 & 14.47 & 1.93 & 49.52 & 23.47 & -26.05 & 0.01 \\
\texttt{deepseek-r1-0528}              & 10.93 & 3.22 & 38.26 & 23.15 & -15.11 & 0.04 \\
\texttt{kimi-k2-thinking}              & 9.97 & 0.96 & 43.41 & 19.29 & -24.12 & 0.03 \\

\midrule
\rowcolor[gray]{0.9}
\multicolumn{7}{l}{\textit{Proprietary LLMs}} \vspace{2pt} \\
\texttt{gemini-2.5-flash}              & 17.04 & \underline{10.61} & 42.44 & 27.65 & -14.79 & 0.01 \\
\texttt{gemini-2.5-pro}                & 20.90 & 9.65 & 50.80 & 30.87 & -19.93 & 0.06 \\
\texttt{gemini-3-pro (high$^*$)}       & 26.37 & \textbf{13.83} & 65.27 & 44.05 & -21.22 & 0.14 \\
\texttt{claude-haiku-4.5 (high$^*$)}   & 17.68 & 8.04 & 49.84 & 28.62 & -21.22 & 0.03 \\
\texttt{claude-sonnet-4.5 (high$^*$)}  & 23.47 & 7.40 & 57.88 & 43.73 & -14.15 & 0.09 \\
\texttt{claude-opus-4.5 (high$^*$)}    & 22.51 & 8.04 & 67.85 & 38.59 & -29.26 & 0.15 \\
\texttt{gpt-4.1-nano}                  & 2.25 & 0.64 & 10.29 & 7.07 & -3.22 & 0.006 \\
\texttt{gpt-4.1-mini}                  & 10.61 & 4.18 & 45.34 & 21.86 & -23.48 & 0.007 \\
\texttt{gpt-4.1}                       & 14.47 & 7.40 & 56.27 & 29.90 & -26.37 & 0.03 \\
\texttt{o4-mini (low)}                 & 22.83 & 8.36 & 56.27 & 36.01 & -20.26 & 0.02 \\
\texttt{o4-mini (medium$^*$)}          & 30.55 & 6.43 & 62.06 & 42.44 & -19.62 & 0.03 \\
\texttt{o4-mini (high)}                & 27.97 & 6.43 & 64.95 & 44.05 & -20.90 & 0.06 \\
\texttt{gpt-5-nano (medium$^*$)}       & 16.08 & 4.82 & 50.16 & 29.26 & -20.90 & 0.005 \\
\texttt{gpt-5-mini (medium$^*$)}       & 42.12 & 0.00 & 63.02 & 44.37 & -18.65 & 0.01 \\
\texttt{gpt-5 (minimal)}               & 30.23 & 0.32 & 56.27 & 37.62 & -18.65 & 0.02 \\
\texttt{gpt-5 (low)}                   & 52.41 & 0.96 & 65.59 & 48.55 & -17.04 & 0.04 \\
\texttt{gpt-5 (medium$^*$)}            & \underline{59.16} & 0.00 & 67.52 & \textbf{57.88} & -9.64 & 0.09 \\
\texttt{gpt-5 (high)}                  & \textbf{61.09} & 0.00 & \underline{68.81} & \underline{57.56} & -11.25 & 0.16 \\
\texttt{gpt-6-luna (medium$^*$)}       & 31.19 & \underline{10.61} & 68.17 & 46.30 & -21.87 & 0.006 \\
\texttt{gpt-6-sol (medium$^*$)}        & 40.84 & 9.32 & \textbf{74.60} & 51.77 & -22.83 & 0.04 \\
\bottomrule
\end{tabular}
}
\caption{Main results of structured disambiguation using various LLMs on ARCS. For reasoning models, the level of reasoning effort is indicated in parentheses; $*$ marks the default configuration. \textbf{Bold} indicates the best performance, while \underline{underline} indicates the second-best. $\Delta$EX denotes the difference between {EX}$_\text{e2e}$ and {EX}$_{\text{disambiguated}}$. Costs are per-task USD price. All open-source LLMs were evaluated using the Fireworks serverless API.}
\label{tab:main_result}
\end{table*}

%% file: sections/5-experiments.tex
\section{Experiments}\label{sec:experiments}

\subsection{Experimental Settings}

We evaluate 23 LLMs on ARCS, including 8 open-source models and 15 proprietary models in both the end-to-end setting and fine-grained setting (see \S\ref{sec:benchmark}). For all applicable models, the temperature is set to zero. For reasoning models except \texttt{o4-mini} and \texttt{gpt-5}, we use the default reasoning effort (denoted by $^*$). For \texttt{o4-mini} and \texttt{gpt-5}, we evaluate all supported reasoning effort levels. Proprietary models are run using their official APIs, while all open-source models are run using the Fireworks serverless API. The exact prompts used for different methods are provided in \autoref{appsec:technical}.

\subsection{Main Results}

\autoref{tab:main_result} presents the performance of structured disambiguation across all 23 LLMs under both settings.

\paragraph{\emph{Text-to-SQL remains a significant challenge in the presence of ambiguity.}}

The best end-to-end performance is achieved by \texttt{gpt-5} at 57.88\%, whereas \texttt{gpt-4.1} achieves only 29.90\%. The recent frontier model, \texttt{gpt-6-sol}, achieves 51.77\% end-to-end execution accuracy, despite its higher SQL generation capabilities (\textbf{EX}$_\text{disambiguated}$). Among open-source models, \texttt{gptoss-120b} performs best, achieving 26.05\%. Other open-source models, including those with substantially larger parameter counts, do not yield better performance. Within the same model family, performance consistently scales from smaller to larger models, as observed in the \texttt{gptoss}, \texttt{claude} and \texttt{gpt} families.

\input{tables/disamb_architecture}

\input{tables/taxonomy_ambrosia}

\paragraph{\emph{Disambiguation is the primary bottleneck on ARCS.}}
A closer examination of the isolated disambiguation and SQL generation metrics reveals that disambiguation constitutes the main performance bottleneck. Across all LLMs, disambiguation scores are substantially lower than SQL generation scores. All models except \texttt{gpt-5} and \texttt{gpt-6-sol} only correctly identify all ambiguity points and interpretations for at most 30\% of the tasks. Precise disambiguation is even more challenging, with no model achieving a Perfect Disambiguation score above 14\%.

\paragraph{\emph{Frontier LLMs exhibit similar text-to-SQL performance but substantially different disambiguation capabilities.}}
For \texttt{gpt-5}, \texttt{claude-opus-4.5}, and \texttt{gemini-3-pro}, which were released around the same time, execution accuracy with ground-truth disambiguation is comparable, ranging from 65.27\% to 68.81\%. In contrast, their {Full Recall} scores span a much wider range, from 22.51\% to 61.09\%. This disparity suggests that disambiguation may be underrepresented in current LLM training objectives.

\paragraph{\emph{Performance scales with reasoning effort on ARCS.}}
For \texttt{gpt-5}, end-to-end execution accuracy improves from 37.62\% to 57.88\% as the reasoning effort increases from minimal to medium, with no further gains observed at higher settings. For \texttt{o4-mini}, end-to-end execution accuracy increases from 36.01\% to 44.05\% as the reasoning effort is increased from low to high. Notably, this scaling behavior is present in both disambiguation and SQL generation, with a more pronounced effect observed for disambiguation.

\subsection{Comparison of Disambiguation Methods}

We compare conversational disambiguation, unstructured interpretation generation~\citep{saparina-lapata-2025-disambiguate}, and structured disambiguation. For conversational disambiguation, we additionally vary user patience, defined as the maximum number of clarification questions the user is willing to answer.

\paragraph{\emph{Structured disambiguation requires significantly lower user effort.}} As shown in \autoref{fig:disamb_arch}, structured disambiguation matches the performance of conversational disambiguation while requiring significantly lower user effort. When user patience is reduced such that the resulting user effort is comparable, conversational disambiguation performs substantially worse. Unstructured interpretation generation exhibits similar performance to the other methods on questions with a single ambiguity point, but degrades markedly in the presence of multiple ambiguity points, consistent with our hypothesis in \S\ref{sec:disambiguation}. We additionally conducted a user study (see \autoref{appsec:user_study}) using the NASA-TLX framework. Consistent with the simulator-based results, participants reported lower mental demand, effort, and frustration with structured disambiguation, while also rating it as more transparent.

\subsection{Comparison with Previous Benchmarks}

In \autoref{tab:taxonomy_ambrosia}, we evaluate structured disambiguation on Ambrosia~\citep{DBLP:conf/nips/SaparinaL24} and ARCS using the same implementation.

\paragraph{\emph{ARCS is taxonomy-proof.}}
When provided with their respective ambiguity taxonomies, \texttt{o4-mini}'s end-to-end execution accuracy increases dramatically from 52\% to 80\% on Ambrosia, but improves only marginally on ARCS. This contrast indicates that ARCS poses fundamentally harder challenges, exhibits greater question diversity and is much less vulnerable to heuristic-based strategies or prompt engineering.

\subsection{Error Analysis}

To verify that the benchmark poses fundamental challenges in disambiguation and SQL generation, we categorize model outputs into a hierarchy of error categories with progressively increasing levels of difficulty, as shown in \autoref{fig:error_distribution}. Although \texttt{qwen3-coder-480b} \citep{yang2025qwen3} produces invalid outputs on 18\% of the tasks—primarily due to inferior instruction-following and tool-calling capabilities—all three models generate executable SQL queries for the majority of inputs. Across all models, the two most prevalent error categories correspond to cases where the generated SQL query is executable but either does not match any ground-truth query or matches a ground-truth query other than the one intended by the user. These results indicate that the dominant sources of error arise from unresolved ambiguities and incorrect SQL formulation.

We further examine the predicted disambiguation and observe several common failure modes. First, \textbf{spurious ambiguity for an unambiguous phrase} is sometimes caused by incorrect relaxation of certain constraints. For example, the phrase ``created pull requests'' clearly maps to the SQL logic \texttt{PullRequestEvent} with \texttt{action = 'opened'}, but the model considers whether to include the filter \texttt{action = 'opened'} as an ambiguity. Second, \textbf{database context is not fully explored} by the model to bound the ambiguity. For example, ``year'' should have only two interpretations when there are only 2022 and 2023 tables in the database, but the model considers ``year'' as an open-ended ambiguity. Third, \textbf{over-fragmentation} occurs when the model predicts multiple interpretation phrases that are semantically identical. For example, ``Q4 of 1996 and full year 1997'' and ``continuous period from 1996-10-01 to 1997-12-31'' are treated as separate interpretations even though they are semantically equivalent.


\input{figures/error_distribution}

\input{figures/result_by_ambig_type}

\subsection{Analysis on Ambiguity Types}

To investigate whether certain types of ambiguity are more challenging than others, we analyze the recall (percentage of ambiguity points identified by the model) across different ambiguity types in \autoref{fig:result_by_ambig_type}. All three models perform poorly at identifying computation ambiguities. In addition, the weaker models, \texttt{qwen3-coder-480b} and \texttt{gpt-4.1}, also struggle to detect syntactic ambiguities compared to semantic ones.

%% file: tables/disamb_architecture.tex
\begin{figure}[ht]
\centering
\begin{minipage}{0.6\textwidth}
    \centering
    \scriptsize
    \captionsetup{type=table}
    \scalebox{1.0}{
    \begin{tabular}{lccc}
    \toprule
    \textbf{Method} & \textbf{EX} (\%) & \textbf{User Effort} & \textbf{Cost} (\$) \\
    \midrule
    \rowcolor[gray]{0.9}
    \multicolumn{4}{l}{\textit{Conversational Disambiguation}} \vspace{2pt} \\
    \texttt{o4-mini (patience = 1)} & 31.51 & \underline{20} & 0.02 \\
    \texttt{o4-mini (patience = 3)} & 39.55 & 27 & 0.02 \\
    \texttt{o4-mini (patience = 5)} & 39.87 & 31 & 0.02 \\
    \texttt{o4-mini} & \textbf{43.73} & 34 & 0.03 \\
    \midrule
    \rowcolor[gray]{0.9}
    \multicolumn{4}{l}{\textit{Unstructured Disambiguation}} \vspace{2pt} \\
    \texttt{o4-mini}                 & 28.94 & 39 & 0.04 \\
    \midrule
    \rowcolor[gray]{0.9}
    \multicolumn{4}{l}{\textit{Structured Disambiguation}} \vspace{2pt} \\
    \texttt{o4-mini}           & \underline{42.44} & \textbf{18} & 0.03 \\
    \bottomrule
    \end{tabular}
    }
    \label{tab:disamb_arch}
\end{minipage}\hfill
\begin{minipage}{0.4\textwidth}
    \centering
    \includegraphics[width=0.9\linewidth]{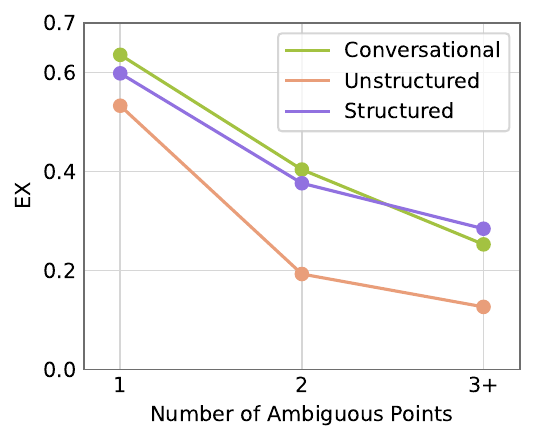}
    \label{fig:disamb_arch_num_ap}
\end{minipage}
\caption{Comparison of conversational, unstructured, and structured disambiguation. Left: Overall Results. The reasoning effort of \texttt{o4-mini} is medium. Patience is the maximum number of clarification questions the user is willing to answer. {User Effort} is defined as 0.1 times the user simulator input tokens plus the number of output tokens. Right: End-to-end EX across varying numbers of ambiguity points.}
\label{fig:disamb_arch}
\end{figure}

%% file: tables/taxonomy_ambrosia.tex
\begin{table*}[ht]
\centering
\scriptsize
\scalebox{1.0}{
\begin{tabular}{lcccc}
\toprule
\textbf{Method} &
\multicolumn{2}{c}{\textbf{w/o Taxonomy}} &
\multicolumn{2}{c}{\textbf{w/ Taxonomy}} \\
\cmidrule(lr){2-3} \cmidrule(lr){4-5}
 & \textbf{ARCS EX} (\%) & \textbf{Ambrosia EX} (\%) &
   \textbf{ARCS EX} (\%)  & \textbf{Ambrosia EX}  (\%) \\
\midrule
\texttt{gpt-4.1} & 29.90 & 48.75 & 28.30 & 63.25 \\
\texttt{o4-mini (medium)} & 42.44 & 52.25 & 46.30 & 80.00 \\
\bottomrule
\end{tabular}
}
\caption{Comparison of end-to-end EX performance on ARCS and Ambrosia, with and without taxonomy as input. Ambrosia is evaluated on a 400-sample subset.}
\label{tab:taxonomy_ambrosia}
\end{table*}

%% file: figures/error_distribution.tex
\begin{figure*}[t]
    \centering
    \includegraphics[width=\linewidth]{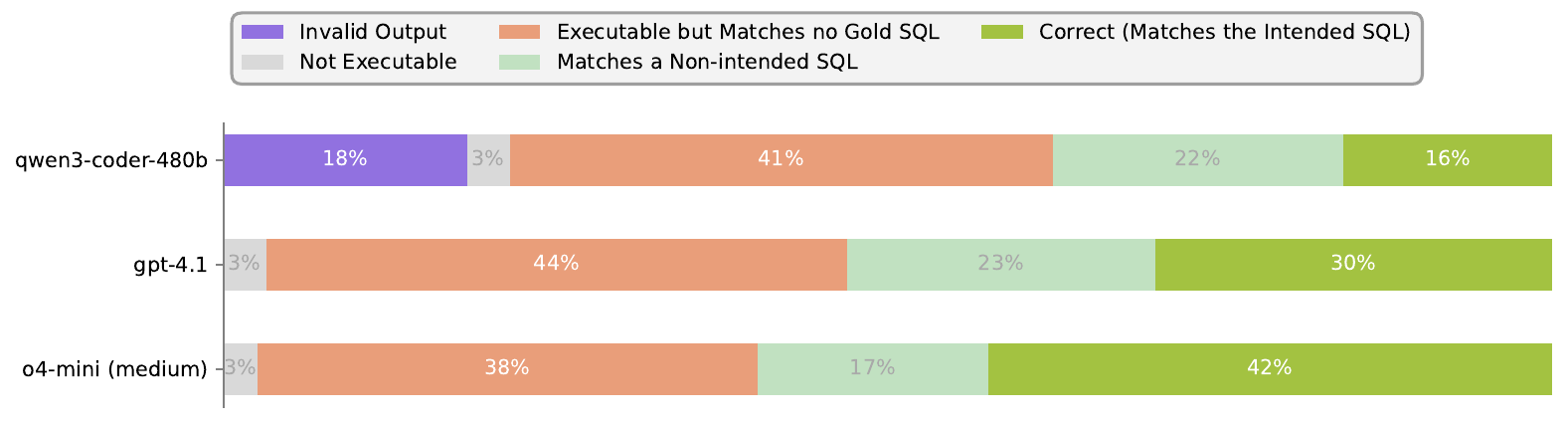}
   \caption{Error distribution of three representative models on the full ARCS dataset. See \autoref{fig:error_distribution_all} for results on the remaining models.}

    \label{fig:error_distribution}
\end{figure*}

%% file: figures/result_by_ambig_type.tex
\begin{figure*}[t]
    \centering
    \includegraphics[width=0.75\linewidth]{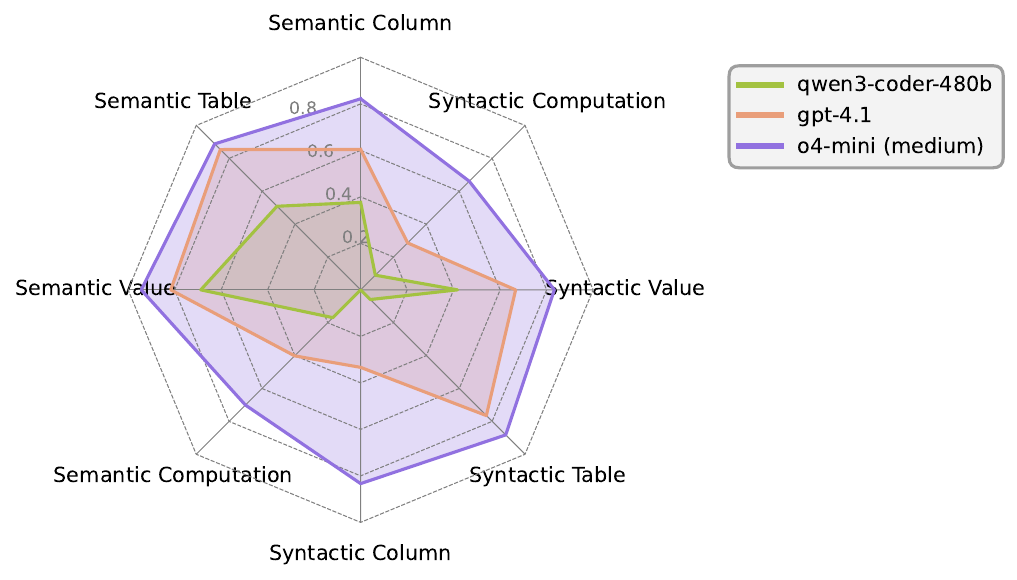}
   \caption{Ambiguity point recall for each ambiguity type.}

    \label{fig:result_by_ambig_type}
\end{figure*}

%% file: sections/7-conclusion.tex
\section{Conclusion}\label{sec:conclusion}

Ambiguity is a fundamental yet largely underexplored challenge in deploying text-to-SQL systems over real-world databases. As a first step toward transparent and efficient disambiguation with minimal user effort, we introduce structured disambiguation, a new interaction paradigm that resolves ambiguity through explicit, constrained interactions. We also present ARCS, the first benchmark of text-to-SQL questions with naturally occurring, unconstrained ambiguity over real-world databases. Our experimental results demonstrate that text-to-SQL remains highly challenging in the presence of ambiguity. We hope that ARCS will catalyze future research on ambiguity-aware text-to-SQL systems and inspire the development of interaction paradigms that better align with realistic user workflows.

%% file: sections/appendix.tex
\input{sections/6-related-work}



\section{Additional Benchmark Details}\label{appsec:benchmark_details}

\subsection{Annotation Details} \label{appsubsec:annotation}

Given that annotating ARCS requires high expertise in both linguistics and databases, the authors conducted all annotations themselves to ensure the highest annotation quality. After the initial generation of candidate questions, the authors carried out the first round of annotation, which involved question selection, ambiguity annotation, and SQL annotation. The work was divided by database.

For each database, annotators first spent approximately two hours reviewing the database schema and gaining familiarity with the entities and relations in the domain. Annotators also read column descriptions when available. For structured columns (e.g., JSON columns), each field within the structure, including its semantics and possible values, was further examined and documented for future reference.

Next, the annotators reviewed all candidate questions generated for each database and selected those that were realistic and genuinely ambiguous. Once the desired number of questions was reached, the annotators proceeded to annotate the ambiguity points and their corresponding interpretations for each selected question. The annotation guidelines were formulated during this first round of annotation and covered the following aspects:

\begin{itemize}[left=10pt]
    \item \textbf{Criteria for classifying ambiguity types}, with particular attention to edge cases. For example, table ambiguity is defined as cases where the only difference in the corresponding SQL queries lies in the table name and its dependent logic.
    \item \textbf{Avoidance of overly vague phrases} such as ``good players''. We focus on ambiguous phrases with clear and well-defined interpretations.
    \item \textbf{Inclusion of only realistic interpretations}. For instance, for the phrase ``March or April 2017'', although the interpretation ``March in any year or April in 2017'' is theoretically plausible, it is excluded because it is not realistic.
    \item \textbf{Avoidance of unanswerable questions}, i.e., questions that cannot be answered or can only be approximately answered using the database.
    \item \textbf{Marking of required columns}. When writing SQL queries, all columns relevant to answering the question are included (e.g., including the score when asked for the student who achieved the highest score), but only explicitly requested columns are marked as required (e.g., the student name). When a question requests a set of entities, the entity name is marked as required if available (e.g., for students), or the identifier otherwise (e.g., for transactions).
    \item \textbf{Ensuring consistency across SQL queries}. Most questions involve annotating multiple SQL queries, and annotators are instructed to maintain consistent syntax whenever possible to reduce the cognitive load during subsequent verification.
    \item \textbf{Ensuring non-empty results}. For each question, at least one annotated SQL query is required to produce non-empty results.
    \item \textbf{Avoidance of time-sensitive expressions} such as ``last year'', ensuring that the intended interpretation and corresponding SQL do not change over time. One exception is the use of terms such as ``young'', for which the intended interpretation can be defined using a fixed date threshold.
    \item \textbf{Annotation conventions} shown in \autoref{tab:dataset_instructions}, such as returning ties for maximum values and not multiplying percentage values by 100.
    \item \textbf{Common ambiguity patterns} (e.g., ``X and Y or Z'' as syntactic computation ambiguity), provided as references to annotators to reduce the likelihood of missing ambiguity points or interpretations.
    \item \textbf{Common SQL snippets} (e.g., SQL snippets for computing age from date of birth), provided as references to annotators to ensure the consistency and correctness of SQL annotations.
\end{itemize}

During annotation, annotators were allowed to revise questions to introduce or remove ambiguity points. They iteratively refined the annotations of ambiguity points, interpretations, and corresponding SQL queries. After the first round of annotation, the annotators conducted a second round by referencing the finalized annotation guidelines. During this process, over 60\% of the questions, especially those annotated earlier, were found to contain annotation errors, such as missing ambiguity points, missing interpretations, or overly vague phrases. 

After the second round of annotation, the annotators performed a final verification pass using the annotation guidelines, during which fewer than 2\% of the questions were identified as containing annotation errors. The annotation process took approximately 2.5 hours per question and resulted in 101 unique questions, which yielded 311 tasks by sampling the intended ambiguity resolutions.

\subsection{Database Statistics}

\input{tables/database_stats}

\autoref{tab:database_stats} shows statistics of the ARCS databases and a comparison with previous benchmarks. The ARCS databases are significantly larger than prior ambiguous text-to-SQL benchmarks~\citep{DBLP:conf/nips/SaparinaL24} and are comparable to the ones in Spider 2.0~\citep{lei2024spider2}.

\subsection{Question Statistics}

\input{figures/benchmark_stats}

\input{figures/ambig_type_distribution}

\autoref{fig:benchmark_stats} summarizes the question statistics, including the distribution of domains, ambiguity points per question, and SQL queries per question. \autoref{fig:ambig_type_distribution} shows the distribution of ambiguity types across all annotated ambiguity points in ARCS. Although semantic ambiguities are the most common in practice, ARCS includes instances of all eight ambiguity types.

\subsection{Global Annotation Conventions}

\input{tables/dataset_instructions}

The complete global annotation conventions described in \S\ref{subsec:annotation} are provided in \autoref{tab:dataset_instructions}.

\subsection{Sample Database Schema}

\input{figures/schema_retails}
\input{figures/schema_basketball}

\autoref{fig:schema_retails} and \autoref{fig:schema_basketball} show the database schemas of the retail and basketball databases.

\subsection{Sample SQL Query}

\input{tables/sample_sql}

\autoref{tab:sample_sql} show a sample SQL query at the median of the length distribution.

\subsection{Data Format}

\input{tables/data_format_sample_1}
\input{tables/data_format_sample_2}

\autoref{tab:data_format_1} and \autoref{tab:data_format_2} illustrates the ARCS data format using a sample task entry.

\section{Additional Results}

\subsection{Error Analysis}

\autoref{fig:error_distribution_all} presents the complete error distribution for all evaluated models.

\input{figures/error_distribution_all}

\clearpage

\section{Additional Technical Details} \label{appsec:technical}

\subsection{Details for Structured Disambiguation} \label{appsubsec:details_structured}

The disambiguation step described in \S\ref{sec:disambiguation} is instantiated using a standard ReAct agent~\citep{yao2022react} equipped with a \texttt{get\_schema} tool that returns the database schema. The disambiguation agent additionally has a \texttt{final\_result} tool, which is used to output the identified structured disambiguation.

The SQL generation step is instantiated using a ReAct agent with access to the \texttt{get\_schema}, \texttt{get\_column\_description}, \texttt{search\_keywords}, and \texttt{run\_query} tools. The SQL generation agent additionally has a \texttt{finish} tool that terminates the agent loop and submits the last executed SQL query as the final output. The system prompts for both agents are shown in \autoref{tab:prompt_structured_disamb} and \autoref{tab:prompt_structured_sql}. The global annotation conventions described in \S\ref{subsec:annotation} are shown in \autoref{tab:dataset_instructions} and are included in the dataset instructions section of the system prompt for both agents.

\input{tables/prompt_structured_disamb}
\input{tables/prompt_structured_sql}
\input{tables/prompt_simple}

\input{tables/prompt_user_1}

\input{tables/prompt_user_2}

\subsection{Details for Conversational Disambiguation}

For conversational disambiguation, we implement a single ReAct agent equipped with a \texttt{ask\_user} tool for asking clarification questions in natural language. The agent also has access to the same tools as the SQL generation agent used in structured disambiguation. The system prompt is shown in \autoref{tab:prompt_simple}.

When user patience is specified, the agent is informed of the maximum number of allowed calls to the \texttt{ask\_user} tool via the system prompt. Once this quota is exhausted, the tool is removed from the list of available tools.



\subsection{Details for Unstructured Interpretation Generation~\citep{saparina-lapata-2025-disambiguate}}

The implementation for unstructured interpretation generation of \citep{saparina-lapata-2025-disambiguate} is similar to our implementation for structured disambiguation: a disambiguation agent first identifies all possible interpretations of the full question, the user then selects the intended one, and finally a SQL generation agent produces the corresponding SQL query. The main difference lies in the output of the disambiguation step: instead of producing a list of ambiguity points, each with its own interpretations, the method generates a list of full-sentence interpretations where each interpretation is a standalone natural-language paraphrase of the entire question.

The original method proposed in \citep{saparina-lapata-2025-disambiguate} cannot handle vagueness. To ensure a fair comparison on ARCS, we adapt the method to support vagueness. Specifically, during the disambiguation phase, we introduce a separate agent to identify ambiguity points corresponding to numeric thresholds. These ambiguity points are also passed to the user simulator for resolution and are later used for SQL generation, in a manner similar to structured disambiguation. System prompts similar to that shown in \autoref{tab:prompt_structured_disamb} are used for disambiguation, and the same prompt shown in \autoref{tab:prompt_structured_sql} is used for SQL generation.

\subsection{Details for User Simulator}

As described in \S\ref{subsec:user_simulator}, we adopt a two-stage user simulator to avoid information leakage. In the first stage, the LLM is prompted to identify the relevant annotated ambiguity points. In the second stage, the LLM is prompted to generate a response conditioned on the identified ambiguity points. The prompts used in the two stages are shown in \autoref{tab:prompt_user_1} and \autoref{tab:prompt_user_2}.

\subsubsection{Details for Evaluation Metrics}

To compute {Full Recall} and {Perfect}, we first use an LLM to align each predicted ambiguity point with the corresponding ground-truth ambiguity point, and then align the interpretations within each matched ambiguity point. We compute precision, recall, and F1 scores at both the ambiguity-point level and the interpretation level. {Full Recall} is set to 1 if both the ambiguity-point recall and the interpretation recall equal 1. {Perfect} is set to 1 if both the ambiguity-point F1 score and the interpretation F1 score equal 1.

Perfect can be viewed as a stricter version of Full Recall that captures precise disambiguation. For a given task, Full Recall = 1 together with EX$_{\text{disambiguated}}$ = 1 typically implies EX$_{\text{e2e}}$ = 1, assuming the user simulator behaves as expected.


\clearpage

\section{User Study}
\label{appsec:user_study}

We conducted a small-scale user study with eight participants to compare conversational and structured disambiguation. None of the participants are authors of this paper. All participants regularly analyze tabular data and use LLM-based tools on a daily or weekly basis. 

\subsection{Study Design}

We developed a terminal interface using tabulaflow\footnote{\url{https://github.com/megagonlabs/tabulaflow}} for both conversational and structured disambiguation. The two systems share most of their implementation code and system prompts to ensure a fair comparison. Both interfaces were polished to behave effectively in their respective interaction settings.

We used the \texttt{retails} database for the study. Participants first completed a practice session to familiarize themselves with the database, the interfaces, and the relevant key bindings. We then used a within-subjects design in which each participant interacted with both systems. Participants asked three questions using one system before switching to the other system for another three questions. The system order was randomized and balanced across the eight participants.

For each system, the first two questions were sampled from benchmark questions, with comparable ambiguity complexity across the two systems to ensure fairness. The third question was created by the participant. All study sessions were screen-recorded.

\subsection{Questionnaire and Results}

After completing the study, participants filled out a questionnaire consisting of six quantitative measures. Four workload measures were adapted from the NASA-TLX framework: Mental Demand, Temporal Demand, Effort, and Frustration. We additionally measured Result Confidence and System Transparency. Each measure was rated on a 1--7 scale. We also collected open-ended feedback about which aspects of the interface contributed most to participants' workload. Table~\ref{tab:user_study} shows the questionnaire items together with the average ratings for the two systems.

\begin{table*}[h]
\centering
\scriptsize
\begin{tabular}{p{0.16\linewidth} p{0.51\linewidth} cc}
\toprule
\textbf{Measure} & \textbf{Question} & \textbf{Conversational} & \textbf{Structured} \\
\midrule
Mental Demand &
How mentally demanding was the task? &
3.50 & 2.75 \\

Temporal Demand &
How hurried or rushed was the pace of the task? &
2.50 & 2.38 \\

Effort &
How hard did you have to work to accomplish your level of performance? &
3.50 & 2.63 \\

Frustration &
How insecure, discouraged, irritated, stressed, and annoyed were you? &
2.88 & 2.00 \\

Result Confidence &
Did you feel confident that the system correctly answered your question? &
5.25 & 5.38 \\

System Transparency &
Did the interface make it clear how the system interpreted your input? &
5.13 & 5.75 \\

Feedback &
What specific aspect of the interface contributed most to your workload? &
-- & -- \\
\bottomrule
\end{tabular}
\caption{User study questionnaire and results. All quantitative measures use a 1--7 scale. For workload measures, lower scores are better; for Result Confidence and System Transparency, higher scores are better.}
\label{tab:user_study}
\end{table*}

Structured disambiguation performed better on all six quantitative measures. The largest differences were observed for Effort and Frustration, both decreasing by 0.88 points, followed by Mental Demand, which decreased by 0.75 points. System Transparency increased by 0.62 points, while Result Confidence increased slightly from 5.25 to 5.38.

Participants' qualitative feedback also highlighted the interpretation panel and preview tables as useful components of the structured interface. Participants reported that these features reduced mental workload and helped them decide among ambiguity options.

%% file: sections/6-related-work.tex
\section{Related Work}\label{sec:related-work}

\subsection{Text-to-SQL Benchmarks with Ambiguous Questions}

Early text-to-SQL benchmarks emphasized generating semantically correct SQL for \emph{unambiguous} questions. WikiSQL~\citep{zhongSeq2SQL2017} and Spider~\citep{Yu&al.18c} were two examples that featured relatively simple schemas, questions, and queries, which differ substantially from real-world use cases involving enterprise data~\cite{bodensohn2024llms}. After the emergence of LLMs, newer benchmarks—BIRD~\citep{li2024can}, BEAVER~\citep{chen2024beaver}, and Spider~2.0~\citep{lei2024spider2}—have sought to incorporate more realistic and complex database schemas and questions. However, these benchmarks still assume a single, unambiguous intent per question, yielding exactly one ground-truth SQL\footnote{Although not explicitly designed to target ambiguity, many of these benchmarks are reported to contain ambiguous questions~\citep{liu2025nl2sql,bhaskar-etal-2023-benchmarking}.
}. This assumption rarely holds in practice, particularly when users lack domain knowledge.


In recent years, several benchmark that explicitly target ambiguity have been proposed.
NoisySP~\citep{wang-etal-2023-know} create column ambiguity and table ambiguity questions based on WikiSQL \citep{zhongSeq2SQL2017} and WTQ~\citep{shi-etal-2020-potential}. Ambiguities are injected by modifying the database, specifically by replacing a column with multiple semantically overlapping ones. AmbiQT~\citep{bhaskar-etal-2023-benchmarking} applies a similar approach on Spider~\cite{Yu&al.18c} but include additional ambiguity types. Unlike NoisySP and AmbiQT,  AMBROSIA~\citep{DBLP:conf/nips/SaparinaL24} focus on three types of ambiguities (scope, attachment, vagueness) along the linguistic dimension. The authors first synthesize small-scale databases (each table contains only 3--5 rows) by using LLMs to instantiate predefined templates for each ambiguity type, and then ask human annotators to write ambiguous questions. These benchmarks suffer from several limitations: (1) each question contains at most a single ambiguity point with two to three interpretations; (2) ambiguity is restricted to a small set of predefined templates; (3) the databases are either synthesized or modified to induce specific ambiguity, limiting their realism; and (4) evaluation primarily focuses on predicting all valid SQL queries without the involvement of a user simulator.


A contemporaneous work, BIRD-INTERACT~\citep{birdinteract2025}, introduces a diverse set of ambiguity types and includes queries beyond SELECT. However, it primarily targets \textit{underspecification}, as its data construction methodology relies on masking existing information or introducing vague terms to make fully-specified queries incomplete. Moreover, its evaluation is entangled; it focuses on end-to-end evaluation with constrained system architectures, and the user simulator is restricted to fixed interaction protocols. In contrast, ARCS provides complete annotations and an evaluation framework for disentangling disambiguation accuracy from SQL generation accuracy.

\subsection{Disambiguation Methods}

Different strategies for handling ambiguity have been explored across NLP tasks, including generating multiple candidate answers \citep{min-etal-2020-ambigqa}, asking clarification questions \citep{lee-etal-2023-asking,Zhang2024ModelingFC}, estimating answer uncertainty \citep{cole-etal-2023-selectively}, iteratively refining interpretations \citep{sun-etal-2023-answering}, and first detecting and then clarifying ambiguous inputs \citep{kim-etal-2024-aligning}. Methods specifically designed for text-to-SQL have also been proposed, including detecting column ambiguity via counterfactuals \citep{wang-etal-2023-know}, special-purpose decoding to produce multiple candidate SQL queries \citep{bhaskar-etal-2023-benchmarking}, iteratively discovering missing interpretations using a separate LLM \citep{saparina-lapata-2025-disambiguate}, and user-in-the-loop clarification pipelines \citep{zhao-etal-2025-sphinteract,dong-etal-2025-practiq,DBLP:conf/icde/ZhangMXZPJ25}.

In this work, we address ambiguity and vagueness by leveraging structured, system-guided feedback (e.g., multiple-choice options) rather than unstructured, user-authored feedback (e.g., free-form text). Although unstructured feedback can capture diverse user intentions, it often makes interpretation more challenging for systems and may introduce additional ambiguities. Moreover, providing natural language feedback can require greater effort from users and can lead to partial or incomplete feedback~\citep{elgohary-etal-2020-speak}. In contrast, structured feedback simplifies response processing and reduces user effort by presenting predefined, unambiguous options.



%% file: tables/database_stats.tex
\begin{table*}[ht]
\centering
\scriptsize
\scalebox{1.0}{
\begin{tabular}{lrrrrrl}
\toprule
\textbf{Benchmark} & \textbf{\# Question} & \textbf{\# DB} & \textbf{\# Table / DB} & \textbf{\# Column / DB} & \textbf{\# Row / DB} & \textbf{Has Ambiguity?} \\
\midrule
BIRD (dev)~\citep{li2024can}   & 1543 & 11  & 6.82   & 72.55   & 357521 & \texttimes \\
Spider 2.0~\citep{lei2024spider2}   & ---  & --- & ---    & ---     & ---       & \texttimes \\
Ambrosia~\citep{DBLP:conf/nips/SaparinaL24}     & 1139 & 846 & 4.93   & 19.25   & 22     & \checkmark  \\
ARCS (ours)        & 311  & 6   & 133.83 & 1205.00 & 1969774   & \checkmark  \\
\bottomrule
\end{tabular}
}
\caption{Database statistics and comparison with previous benchmarks.}
\label{tab:database_stats}
\end{table*}

%% file: figures/benchmark_stats.tex
\begin{figure}[htbp]
    \centering
    \begin{subfigure}{0.32\textwidth}
        \centering
        \includegraphics[width=\linewidth]{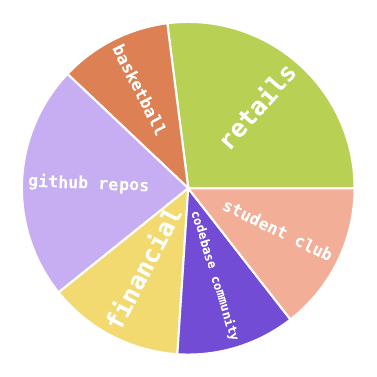}
    \end{subfigure}
    \hfill
    \begin{subfigure}{0.32\textwidth}
        \centering
        \includegraphics[width=\linewidth]{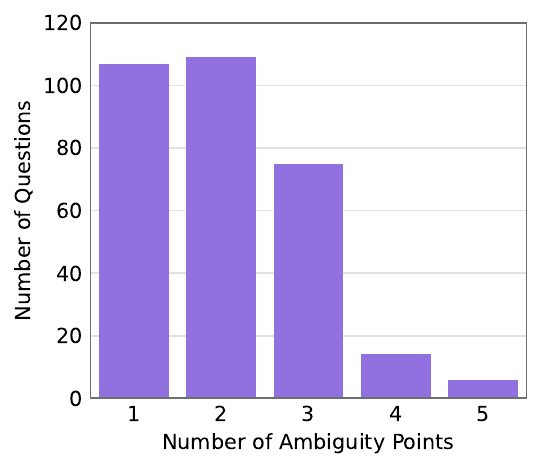}
    \end{subfigure}
    \hfill
    \begin{subfigure}{0.32\textwidth}
        \centering
        \includegraphics[width=\linewidth]{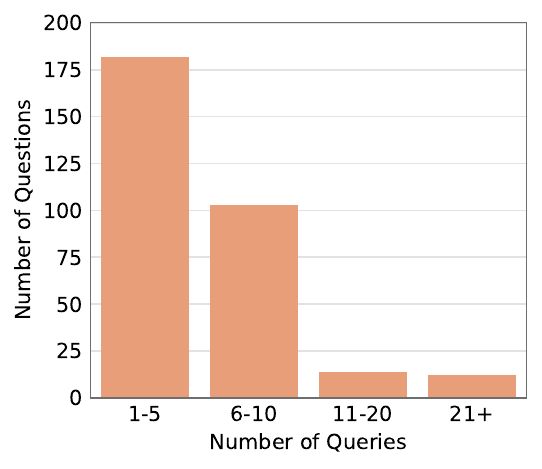}
    \end{subfigure}
    \caption{Benchmark statistics: (left) domain distribution, (middle) distribution of ambiguity points per question, and (right) distribution of SQL queries per question.}
    \label{fig:benchmark_stats}
\end{figure}

%% file: figures/ambig_type_distribution.tex
\begin{figure*}[t]
    \centering
    \includegraphics[width=0.6\linewidth]{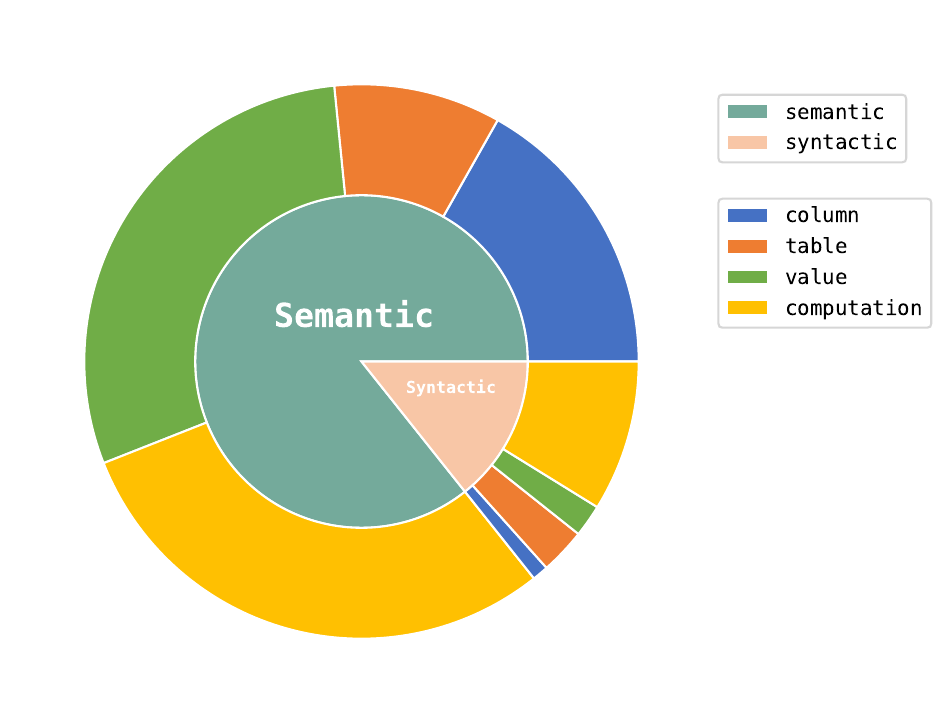}
   \caption{Distribution of ambiguity types across all annotated ambiguity points in ARCS.}
    \label{fig:ambig_type_distribution}
\end{figure*}

%% file: tables/dataset_instructions.tex
\begin{table*}[h]
\vspace{-5pt}
\centering
\begin{tcolorbox}[title={\scriptsize Global Annotation Conventions of ARCS}, boxrule=0.8pt, colframe=gray, boxsep=2pt, left=2pt, right=2pt, top=2pt, bottom=2pt]
            \begin{lstlisting}[style=TinyJSON]
- Follow these requirements when writing SQL. When disambiguating, do not consider these as ambiguities:
  - If the question asks for a list of objects, return their names if available (e.g. for students), otherwise return their IDs (e.g. for transactions).
  - You may include additional relevant columns that are mentioned in the question, even if they are not explicitly requested in the output.
  - Do not concatenate columns in the results unless explicitly requested.
  - For percentage values, don't multiply by 100.
  - Rounding is not needed for numerical values.
  - If the question asks for the object that achieves the maximum/minimum value, if there is a tie, return all tied objects.
=== END OF EXAMPLE ===
\end{lstlisting}
\end{tcolorbox}
\caption{Global annotation conventions of ARCS. These instructions are provided to the system during both disambiguation and SQL generation.}
\label{tab:dataset_instructions}
\end{table*}

%% file: figures/schema_retails.tex
\begin{figure*}[t]
    \centering
    \includegraphics[width=\linewidth]{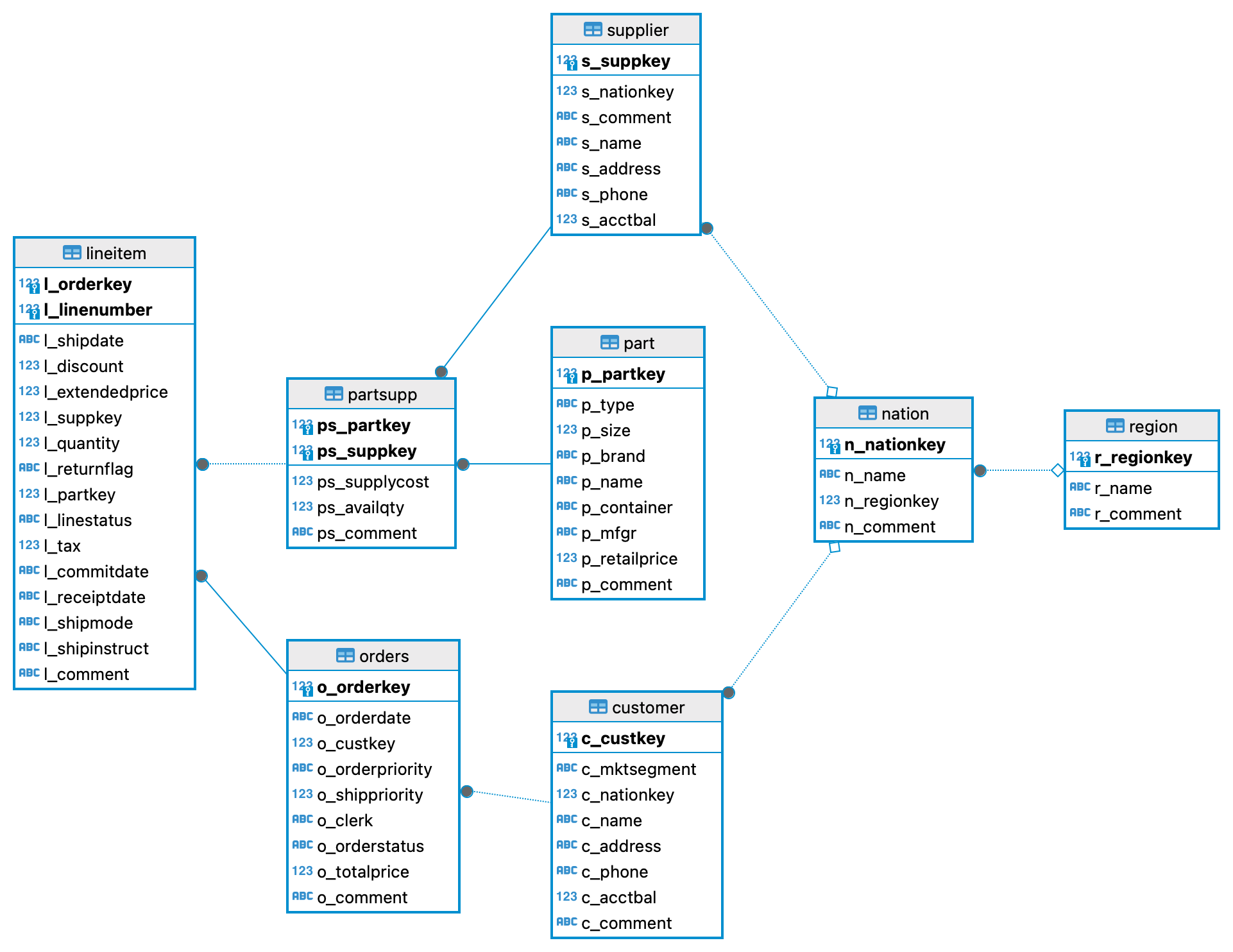}
   \caption{Schema of the retails database.}

    \label{fig:schema_retails}
\end{figure*}

%% file: figures/schema_basketball.tex
\begin{figure*}[t]
    \centering
    \includegraphics[width=\linewidth]{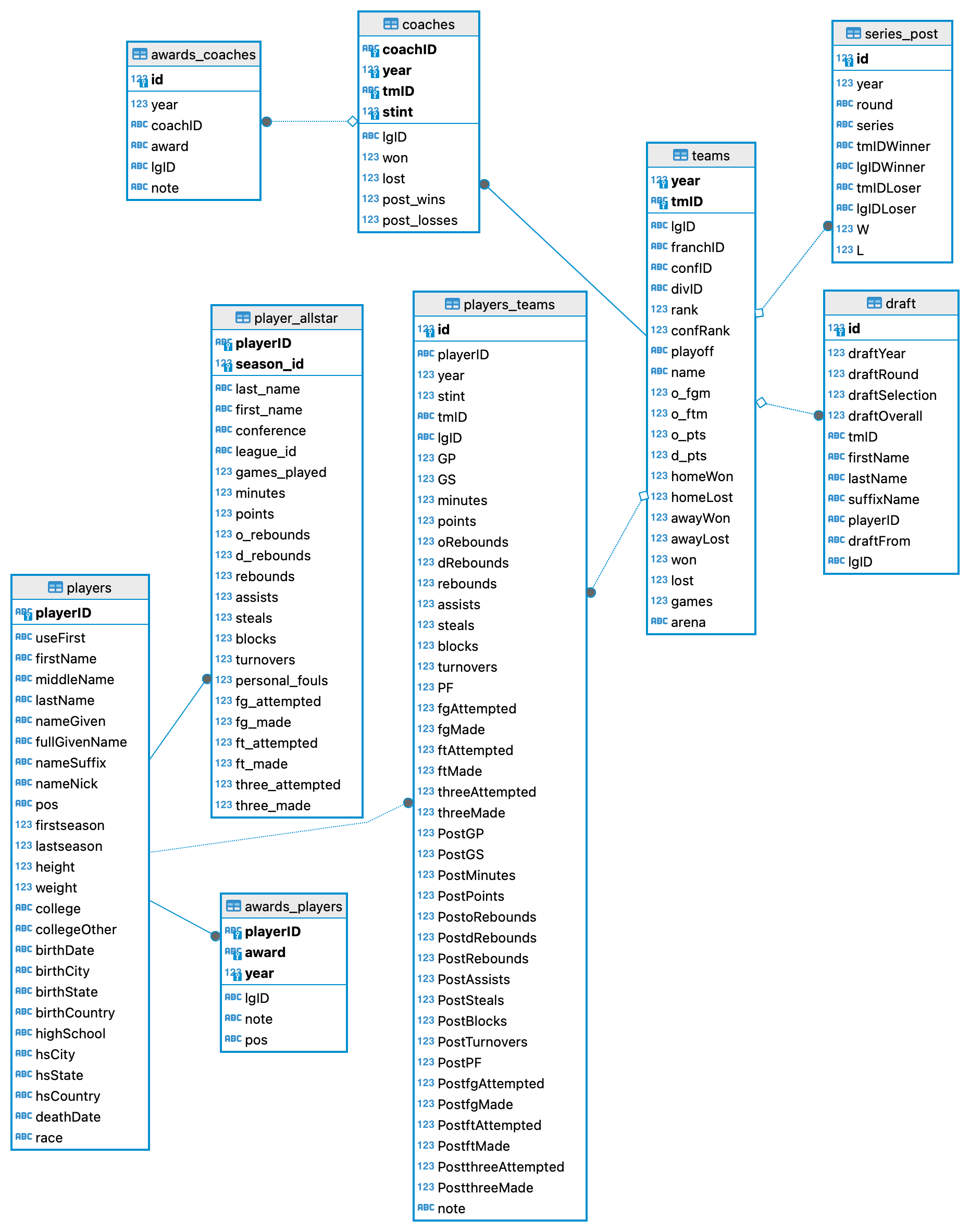}
   \caption{Schema of the basketball database.}

    \label{fig:schema_basketball}
\end{figure*}

%% file: tables/sample_sql.tex
\begin{table*}[h]
\vspace{-5pt}
\centering
\begin{tcolorbox}[title={\scriptsize Sample SQL Query in ARCS}, boxrule=0.8pt, colframe=gray, boxsep=2pt, left=2pt, right=2pt, top=2pt, bottom=2pt]
            \begin{lstlisting}[style=TinySQL]
WITH manufacturer_profit AS (
    SELECT
        p.p_mfgr,
        SUM(
            l.l_extendedprice * (1 - l.l_discount)
            - ps.ps_supplycost * l.l_quantity
        ) AS profit
    FROM lineitem l
    JOIN part p
        ON l.l_partkey = p.p_partkey
    JOIN partsupp ps
        ON l.l_partkey = ps.ps_partkey
       AND l.l_suppkey = ps.ps_suppkey
    WHERE l.l_commitdate >= DATE '1994-10-01'
      AND l.l_commitdate <  DATE '1995-01-01'
    GROUP BY
        p.p_mfgr
)
SELECT
    p_mfgr,
    profit
FROM manufacturer_profit
WHERE profit = (
    SELECT MAX(profit)
    FROM manufacturer_profit
);
\end{lstlisting}
\end{tcolorbox}
\caption{A sample SQL query at the median of the length distribution in ARCS.}
\label{tab:sample_sql}
\end{table*}

%% file: tables/data_format_sample_1.tex
\begin{table*}[h]
\vspace{-5pt}
\centering
\begin{tcolorbox}[title={\scriptsize ARCS Data Format (Part 1 of 2)}, boxrule=0.8pt, colframe=gray, boxsep=2pt, left=2pt, right=2pt, top=2pt, bottom=2pt]
            \begin{lstlisting}[style=TinyJSON]
{
  "qid": "001",
  "task_type": "ambig",
  "has_intended_resolution": true,
  "language": "SQLite",
  "db": "retails",
  "question": "Report the total revenue for each nation in 1995.",
  "gold_ambiguity_points": [
    {
      "id": "A",
      "phrase": "total revenue",
      "type": "finite",
      "ambiguity_type": "semantic_computation",
      "interpretations": [
        "before discount (Gross revenue)",
        "after discount (Net revenue)"
      ],
      "intended_interpretation_idx": 0
    },
    {
      "id": "B",
      "phrase": "total revenue",
      "type": "finite",
      "ambiguity_type": "semantic_computation",
      "interpretations": [
        "include returned items",
        "exclude returned items"
      ],
      "intended_interpretation_idx": 0
    },
    {
      "id": "C",
      "phrase": "for each nation",
      "type": "finite",
      "ambiguity_type": "semantic_table",
      "interpretations": [
        "for each customer nation",
        "for each supplier nation"
      ],
      "intended_interpretation_idx": 1
    },
    {
      "id": "D",
      "phrase": "in 1995",
      "type": "finite",
      "ambiguity_type": "semantic_column",
      "interpretations": [
        "order date in 1995",
        "commitment date in 1995",
        "shipment date in 1995",
        "receipt date in 1995"
      ],
      "intended_interpretation_idx": 3
    }
  ],
  "gold_queries": [
    {
      "id": "GQRY-A.0-B.0-C.0-D.0",
      "query": "WITH revenue AS (\nSELECT n.n_nationkey,\nn.n_name,\nCOALESCE(SUM(l.l_extendedprice), 0) AS total_revenue\nFROM lineitem l\nJOIN orders o ON o.o_orderkey = l.l_orderkey\nJOIN customer c ON o.o_custkey = c.c_custkey\nJOIN nation n ON c.c_nationkey = n.n_nationkey\nWHERE strftime('%Y', o.o_orderdate) = '1995'\nGROUP BY n.n_nationkey, n.n_name\n)\nSELECT n.n_nationkey,\nn.n_name,\nCOALESCE(r.total_revenue, 0) AS total_revenue\nFROM nation n\nLEFT JOIN revenue r ON r.n_nationkey = n.n_nationkey;",
      ... [OMITTED] ...
    }
    {
      "id": "GQRY-A.0-B.0-C.1-D.3",
      "query": "WITH revenue AS (\nSELECT n.n_nationkey,\nn.n_name,\nCOALESCE(SUM(l.l_extendedprice), 0) AS total_revenue\nFROM lineitem AS l\nJOIN supplier AS s ON s.s_suppkey = l.l_suppkey\nJOIN nation AS n ON s.s_nationkey = n.n_nationkey\nWHERE strftime('%Y', l.l_receiptdate) = '1995'\nGROUP BY n.n_nationkey, n.n_name\n)\nSELECT n.n_nationkey,\nn.n_name,\nCOALESCE(r.total_revenue, 0) AS total_revenue\nFROM nation n\nLEFT JOIN revenue r ON r.n_nationkey = n.n_nationkey;",
      "parameter_names": [],
      "parameter_values": {},
      "exec_result": {
        "df": {
          "schema": {
            "dtypes": {
              "n_nationkey": "int64",
              "n_name": "object",
              "total_revenue": "float64"
            }
          },
          ... [SEE PART 2] ...
\end{lstlisting}
\end{tcolorbox}
\caption{Illustration of the ARCS data format using a sample task entry (Part 1 of 2).}
\label{tab:data_format_1}
\end{table*}

%% file: tables/data_format_sample_2.tex
\begin{table*}[h]
\vspace{-5pt}
\centering
\begin{tcolorbox}[title={\scriptsize ARCS Data Format (Part 2 of 2)}, boxrule=0.8pt, colframe=gray, boxsep=2pt, left=2pt, right=2pt, top=2pt, bottom=2pt]
            \begin{lstlisting}[style=TinyJSON]
... [SEE PART 1] ...
          "data": [
            {
              "n_nationkey": 0,
              "n_name": "ALGERIA",
              "total_revenue": 1451887772.1300025
            },
            {
              "n_nationkey": 1,
              "n_name": "ARGENTINA",
              "total_revenue": 1380058942.979993
            },
            ... [OMITTED] ...
          ]
        },
        "df_is_truncated": false,
        "error": null,
        "latency_seconds": 3.4822895526885986
      },
      "other_exec_results": [],
      "required_columns": [
        1,
        2
      ],
      "required_sorted": false
    },
    ... [OMITTED] ...
    {
      "id": "GQRY-A.1-B.1-C.1-D.3",
      "query": ... [OMITTED] ...
    }
  ]
  "gold_intended_query_id": "GQRY-A.0-B.0-C.1-D.3",
  "extra_info": {}
}
\end{lstlisting}
\end{tcolorbox}
\caption{Illustration of the ARCS data format using a sample task entry (Part 2 of 2).}
\label{tab:data_format_2}
\end{table*}

%% file: figures/error_distribution_all.tex
\begin{figure*}[t]
    \centering
    \includegraphics[width=\linewidth]{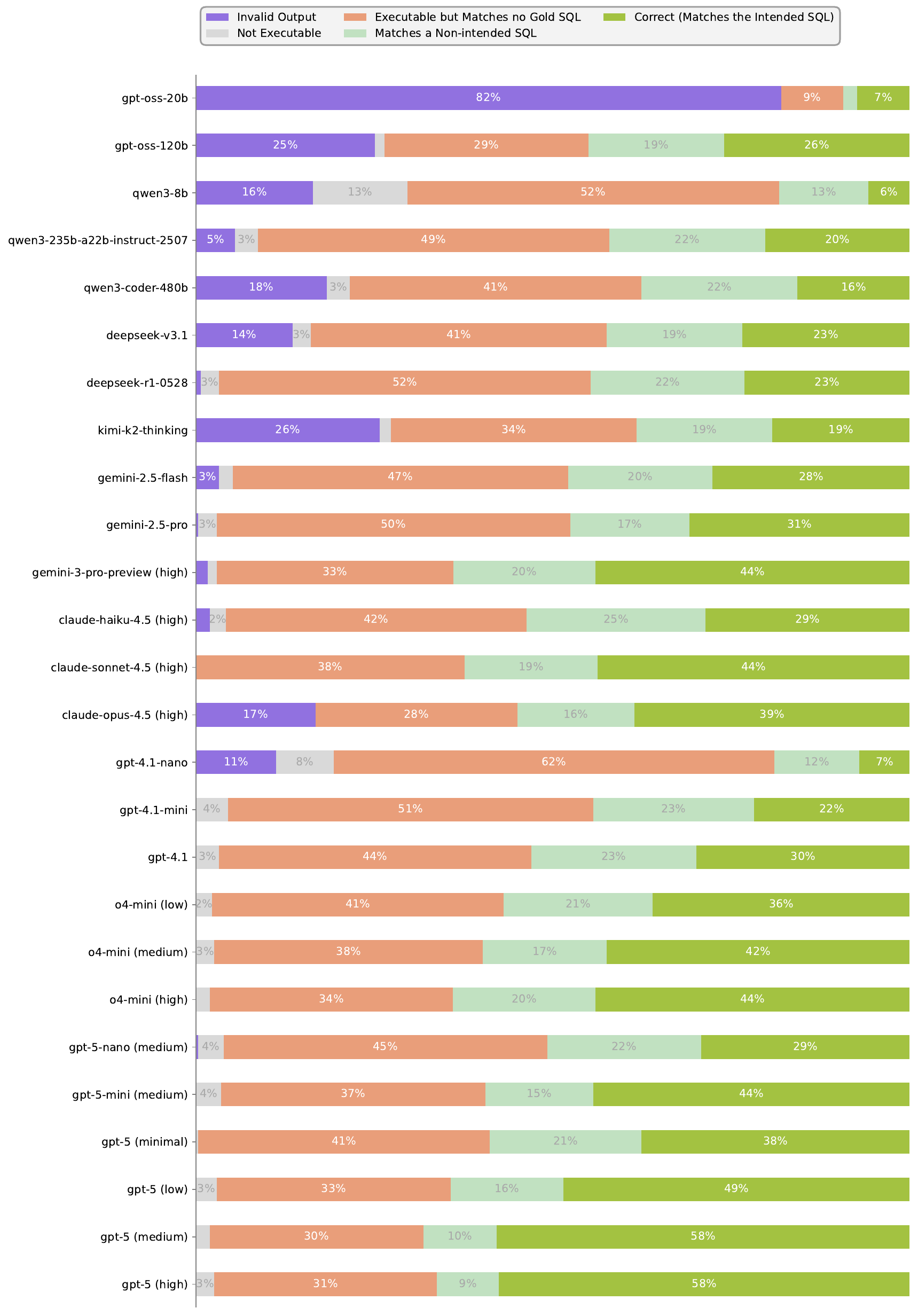}
   \caption{Error distribution of all evaluated models on the full ARCS dataset.}
    \label{fig:error_distribution_all}
\end{figure*}

%% file: tables/prompt_structured_disamb.tex
\begin{table*}[h]
\vspace{-5pt}
\centering
\begin{tcolorbox}[title={\scriptsize Disambiguation Agent System Prompt for Structured Disambiguation}, boxrule=0.8pt, colframe=gray, boxsep=2pt, left=2pt, right=2pt, top=2pt, bottom=2pt]
            \begin{lstlisting}[style=TinyJSON]
You are a helpful AI database expert that can disambiguate questions about a {{language}} database.
The question has one or multiple ambiguity points and you will need to output the list of ALL ambiguity points in the question. Try to be comprehensive.

- For phrases where the number of interpretations is finite, put the list of all possible disambiguated interpretations in the `finite_ambiguity_points` field.
  - There should be at least two interpretations for a phrase to be ambiguous.
  - Each interpretation should be unambiguous and answerable by the database.
  - Each interpretation should be exclusive - only one can apply at a time.
  - If there are multiple dimensions of ambiguity for a phrase, split them into multiple ambiguity points. You can have multiple ambiguity points for one phrase.
  - Do not add number index prefixes to the interpretations.
- For phrases with threshold-like ambiguities (e.g. "tall", "young", etc.), put them in the `parameter_ambiguity_points` field.
  - parameter_sample_operators is a list of valid operators that can be used in <expr> <operator> :<parameter_name>.
  - parameter_sample_values is a list of sample values ordered from least strict to most strict

{% if dataset_instructions %}=== START OF DATASET INSTRUCTIONS ===
{{dataset_instructions}}
=== END OF DATASET INSTRUCTIONS ==={% endif %}

=== START OF EXAMPLE ===
User: List all ambiguity points: List all students with high GPA from NY.

Assistant:
<function name="get_schema">
</function>

Tool:
```
=== TABLE: student (50 rows) ===
- id: INTEGER (e.g. 1) [PK]
- name: TEXT (e.g. "John Doe")
- gpa: FLOAT (e.g. 3.5)
- city: TEXT (e.g. "Los Angeles")
- state: TEXT (e.g. "CA")
=== END OF TABLE ===
```

Assistant:
<function name="final_result">
<arg name="finite_ambiguity_points">
[
  {
    "phrase": "NY",
    "interpretations": [
      "New York City",
      "New York State"
    ]
  }
]
</arg>
<arg name="parameter_ambiguity_points">
[
  { 
    "phrase": "high GPA",
    "parameter_name": "gpa_threshold",
    "parameter_dtype": "float",
    "parameter_description": "GPA threshold to be considered high",
    "parameter_sample_operators": [">", ">="],
    "parameter_sample_values": [3.5, 4.0]
  }
]
</arg>
</function>
=== END OF EXAMPLE ===
\end{lstlisting}
\end{tcolorbox}
\caption{System prompt for the disambiguation agent used in structured disambiguation. The disambiguation agent has access to the \texttt{get\_schema} and \texttt{final\_result} tool.} \label{tab:prompt_structured_disamb}
\end{table*}

%% file: tables/prompt_structured_sql.tex
\begin{table*}[h]
\vspace{-5pt}
\centering
\begin{tcolorbox}[title={\scriptsize SQL Generation Agent System Prompt for Structured Disambiguation}, boxrule=0.8pt, colframe=gray, boxsep=2pt, left=2pt, right=2pt, top=2pt, bottom=2pt]
            \begin{lstlisting}[style=TinyJSON]
You are a helpful AI database expert that can translate natural language questions into {{language}} queries by leveraging the given tools.

- Do not attempt to resolve additional ambiguities with the user. Proceed with the provided information.
- You need to execute the query at least once before finishing. The last executed query will be the final output.
- Ensure the query accurately reflects the original question without adding or omitting any conditions.
- Adhere strictly to the given database schema when constructing queries.
- If you use any of the provided parameters,
  - write a parameterized query with placeholders in the format of `<expr> <operator> :<param_name>`
  - pass in the parameters in the `parameters` field when using the `run_query` tool
- If you think the last executed query is correct, call the `finish` tool with no arguments. Do not output text.

{% if dataset_instructions %}=== START OF DATASET INSTRUCTIONS ===
{{dataset_instructions}}
=== END OF DATASET INSTRUCTIONS ==={% endif %}
\end{lstlisting}
\end{tcolorbox}

\caption{System prompt for the SQL generation agent used in structured disambiguation. The agent has access to the \texttt{get\_schema}, \texttt{get\_column\_description}, \texttt{search\_keywords}, \texttt{run\_query} and \texttt{finish} tools.}

 \label{tab:prompt_structured_sql}
\end{table*}

%% file: tables/prompt_simple.tex
\begin{table*}[h]
\vspace{-5pt}
\centering
\begin{tcolorbox}[title={\scriptsize Agent System Prompt for Conversational Disambiguation}, boxrule=0.8pt, colframe=gray, boxsep=2pt, left=2pt, right=2pt, top=2pt, bottom=2pt]
            \begin{lstlisting}[style=TinyJSON]
You are MintQ agent, a helpful AI database expert that can translate natural language questions into {{language}} queries by leveraging the given tools.

- The question has one or multiple ambiguity points and you will need to ask the user to resolve the ambiguity.
- Do not repeat the question if user refused to answer it.
{% if user_patience -%}
- You have in total {{user_patience}} attempts to call `ask_user` tool to resolve the ambiguity.
  - Only ask one question about one ambiguity point at a time. Try to be comprehensive of all possible ambiguities.
- If the `ask_user` tool is not provided, it means you have used up all your attempts to resolve the ambiguity.
  - Do not attempt to resolve additional ambiguities with the user. Proceed to write the query with the provided information.
{% else -%}
- You are allowed to ask multiple times but only ask one question about one ambiguity point at a time. Try to be comprehensive of all possible ambiguities.
{% endif -%}
- You need to execute the query at least once before finishing. The last executed query will be the final output.
- Ensure the query accurately reflects the original question without adding or omitting any conditions.
- Adhere strictly to the given database schema when constructing queries.
- If you think the last executed query is correct, call the `finish` tool with no arguments. Do not output text.

{% if dataset_instructions -%}
=== START OF DATASET INSTRUCTIONS ===
{{dataset_instructions}}
=== END OF DATASET INSTRUCTIONS ===
{% endif -%}
\end{lstlisting}
\end{tcolorbox}
\caption{System prompt for the agent used in conversational disambiguation. The agent has access to the \texttt{ask\_user}, \texttt{get\_schema}, \texttt{get\_column\_description}, \texttt{search\_keywords}, \texttt{run\_query} and \texttt{finish} tools.}
\label{tab:prompt_simple}
\end{table*}

%% file: tables/prompt_user_1.tex
\begin{table*}[h]
\vspace{-5pt}
\centering
\begin{tcolorbox}[title={\scriptsize User Simulator Stage 1 Prompt}, boxrule=0.8pt, colframe=gray, boxsep=2pt, left=2pt, right=2pt, top=2pt, bottom=2pt]
            \begin{lstlisting}[style=TinyJSON]
You are a data analyst trying to solve the following task: {{task}}
The task has the following ambiguity points:
{% for ap in ambig_points %}
[{{ap.id}}] "{{ap.phrase}}"
  {% if ap.all_interpretations -%}
  Possible interpretations:
  {% for interpretation in ap.all_interpretations -%}
  - {{interpretation}}
  {% endfor -%}
  {% endif -%}
  The intended interpretation is: "{{ap.intended_interpretation}}".
{% endfor %}

You will be asked a question regarding the possible ambiguities in the task, you need to identify the relevant ambiguity point id:
- If there is no matching ambiguity point, return null (or an empty list, depending on the output specification).
- If multiple questions are asked, only identify the relevant ambiguity point for the first question and ignore the rest.
- If the question is not related to ambiguity clarification, return null (or an empty list).
\end{lstlisting}
\end{tcolorbox}
\caption{Prompt used in stage 1 of the user simulator.}
\label{tab:prompt_user_1}
\end{table*}

%% file: tables/prompt_user_2.tex
\begin{table*}[h]
\vspace{-5pt}
\centering
\begin{tcolorbox}[title={\scriptsize User Simulator Stage 2 Prompt}, boxrule=0.8pt, colframe=gray, boxsep=2pt, left=2pt, right=2pt, top=2pt, bottom=2pt]
            \begin{lstlisting}[style=TinyJSON]
You are a data analyst trying to solve the following task: {{task}}
The task has the following ambiguity points:
{% for ap in ambig_points %}
[{{ap.id}}] "{{ap.phrase}}"
  {% if ap.all_interpretations -%}
  Possible interpretations:
  {% for interpretation in ap.all_interpretations -%}
  - {{interpretation}}
  {% endfor -%}
  {% endif -%}
  The intended interpretation is: "{{ap.intended_interpretation}}".
{% endfor %}

You will be asked a question regarding the possible ambiguities in the task, and you are responsible for providing clarifications using the above information.

For "free_text" questions, you need to provide a free-text answer:
- If multiple questions are asked, only answer the first question and say "Please only ask one question at a time."
- Only answer what you are asked using only the information provided above. Do not include additional information.
- Do not make assumptions beyond the provided information.
- You should not ask questions.
- Your answer should be grammatical and linguistically diverse.

For "multiple_choice" questions, you need to select from the given options:
- If none of the options are correct, or cannot be determined using the provided information, set answer number to null. 

For "value" questions, you need to select a value as well as an operator from the given options:
- The data type of the value should be consistent with the one specified in the question.
- If no valid value or no valid operator is correct, or cannot be determined using the provided information, set both fields to null.
\end{lstlisting}
\end{tcolorbox}
\caption{Prompt used in stage 2 of the user simulator.}
\label{tab:prompt_user_2}
\end{table*}